\documentclass[10pt,twocolumn,letterpaper]{article}

\usepackage{iccv}
\usepackage{times}
\usepackage{epsfig}
\usepackage{graphicx}
\usepackage{amsmath}
\usepackage{amssymb}
\usepackage{amsthm}

\usepackage{bbm} 
\usepackage{mathtools}
\usepackage{makecell}
\usepackage{multirow}

\def\x{{\mathbf x}}

\def\D{{\cal D}}
\def\B{{\cal B}}
\def\P{{\cal P}}
\def\X{{\cal X}}
\def\Y{{\cal Y}}
\def\Z{{\cal Z}}

\def\E{{\mathbb E}}

\DeclareMathOperator*{\argmax}{arg\,max}

\newtheorem{theorem}{Theorem}

\newtheorem{proposition}{Proposition}

\usepackage[pagebackref=true,breaklinks=true,letterpaper=true,colorlinks,bookmarks=false]{hyperref}

\iccvfinalcopy % *** Uncomment this line for the final submission

\def\iccvPaperID{8784} % *** Enter the ICCV Paper ID here

\ificcvfinal\fi

\usepackage{fancyhdr}
\begin{document}

%%%%%%%%% TITLE
\title{Unsupervised Accuracy Estimation of Deep Visual Models using Domain-Adaptive Adversarial Perturbation without Source Samples}

\author{%
    JoonHo Lee$^{1}$, Jae Oh Woo$^{2}$, Hankyu Moon$^{2}$ and Kwonho Lee$^{1}$\\
    $^{1}$Samsung SDS, $^{2}$Samsung SDS America\\
    {\tt\small\{joonholee,jaeoh.w,hankyu.m,kwonho81.lee\}@samsung.com}
}
\maketitle
\ificcvfinal\thispagestyle{empty}\fi

%%%%%%%%% ABSTRACT
\begin{abstract}
Deploying deep visual models can lead to performance drops due to the discrepancies between source and target distributions. 
Several approaches leverage labeled source data to estimate target domain accuracy, but accessing labeled source data is often prohibitively difficult due to data confidentiality or resource limitations on serving devices.
Our work proposes a new framework to estimate model accuracy on unlabeled target data without access to source data. 
We investigate the feasibility of using pseudo-labels for accuracy estimation and evolve this idea into adopting recent advances in source-free domain adaptation algorithms.
Our approach measures the disagreement rate between the source hypothesis and the target pseudo-labeling function, adapted from the source hypothesis. 
We mitigate the impact of erroneous pseudo-labels that may arise due to a high ideal joint hypothesis risk by employing adaptive adversarial perturbation on the input of the target model.
Our proposed source-free framework effectively addresses the challenging distribution shift scenarios and outperforms existing methods requiring source data and labels for training.

\end{abstract}

\vspace{-1ex}

%%%%%%%%% BODY TEXT
% ----------------------------------------------------------------------------------
\section{Introduction}
\label{sec:intro}

When deep learning models are deployed for target applications, it is common to encounter a degradation in the accuracy of the models. 
This degradation is typically caused by a distributional discrepancy between the {\it source} domain on which models were trained and the {\it target} domain where they are being applied.
To ensure reliable deployment, it is essential to continually monitor the performance of the deployed model on target data. 
However, since target labels are usually not immediately available, this necessity poses a challenging problem, unsupervised accuracy estimation (UAE) for models.

A few approaches have been developed to predict the target domain accuracy on unlabeled target data assuming that labeled source data can be freely accessed 
\cite{ICML2021_Deng, CVPR2021_Deng, ICCV2021_DoC}.
Nakkiran and Bansal \cite{Nakkiran_Bansal} have observed that the disagreement rate between two separately trained source models exhibiting a small test error of $\epsilon$, is remarkably similar to $\epsilon$ across a range of models.
Subsequent studies have expanded this finding \cite{NEURIPS2021_RM, ICLR2022_GDE}. 
They discovered that the expected disagreement rate of multiple pairs of models or from iterative self-training with the ensemble can effectively estimate model accuracy when tested on previously unseen data. 
Despite the noteworthy outcomes, these methods have a crucial limitation: they are solely applicable to in-distribution test data as indicated in \cite{ICLR2022_GDE}.
In the most realistic scenarios, source and target distributions are supposed to have discrepancies for various reasons.
Insufficient training data is a common cause of generalization issues, which is particularly relevant in scenarios where knowledge from synthetic simulation data is applied in real-world applications, such as autonomous driving and robotic manipulations. 
Also, distribution shifts can occur when input is distorted during sensing and pre-processing.

Two recent approaches \cite{NEURIPS2021_RM, ICML2020_DIR} have adopted unsupervised domain adaptation (UDA) methods, specifically, DANN \cite{DANN} to tackle out-of-distribution UAE.
These methods train supplementary models that learn domain-invariant representations for both the source and the target distributions to identify the maximum discrepancy between the adapted hypotheses and the source hypothesis \cite{ICML2020_DIR} or to employ an ensemble of the adapted hypotheses for iterative self-training \cite{NEURIPS2021_RM}. 
However, focusing on learning domain-invariant representations through DANN can lead to suboptimal outcomes when significant label distribution mismatches occur, increasing the lower bound of the ideal joint hypothesis risk \cite{ICML2019_Zhao}. As a consequence, the estimation performance of these methods is negatively affected.

%------------------------------------------------------------------------
\begin{figure*}[!t]
  \centering
  \includegraphics[width=0.72\textwidth]{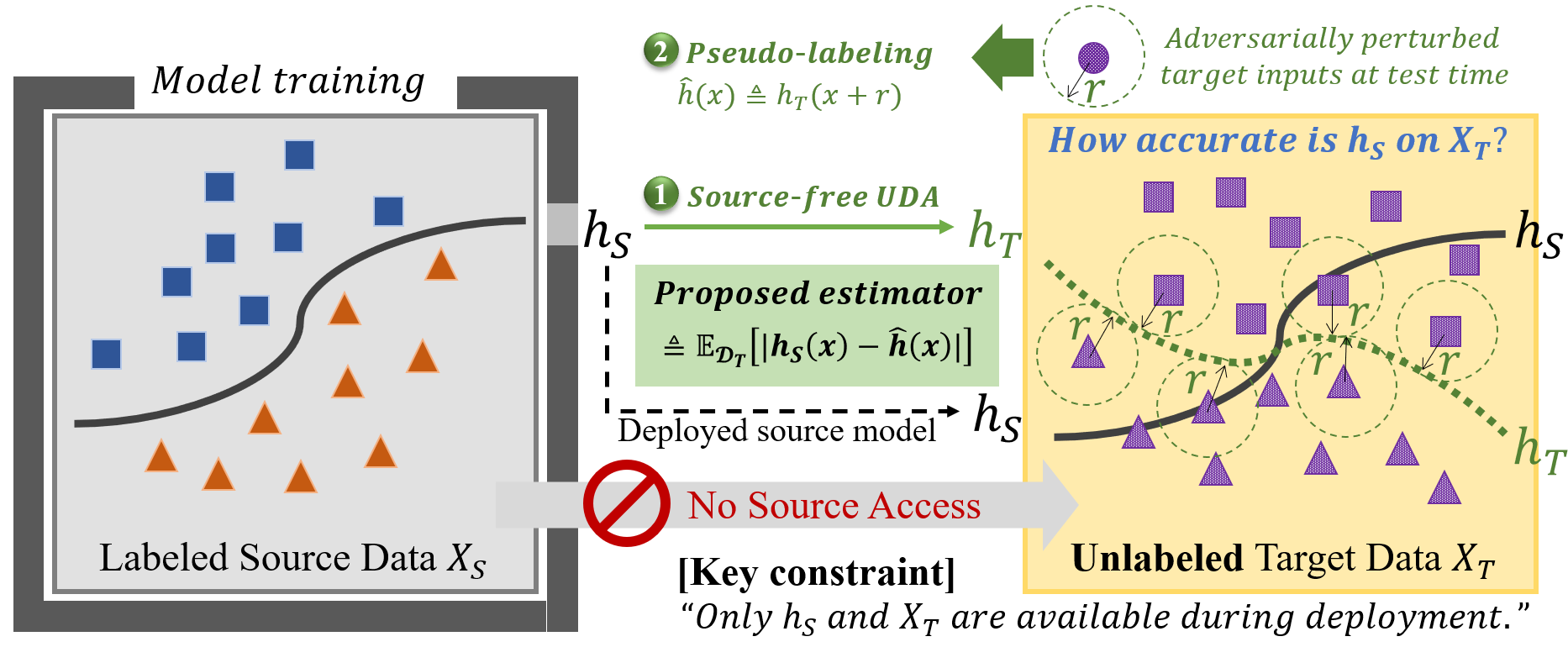}
%\vspace{-1.4ex}
\caption{
Schematic illustration of the proposed SF-DAP (Source-Free Domain-Adaptive Pseudo-labeling) framework. SF-DAP employs source-free UDA and introduces target-adaptive VAPs to tackle the UAE problem without accessing source samples.
}
\label{fig:concept}
\end{figure*}
%------------------------------------------------------------------------

Furthermore, all existing approaches require access to source samples, except for a few naive techniques like AC \cite{ICCV2021_DoC}. 
However, obtaining labeled source data, which are used to train the model of interest, is often infeasible in practice due to concerns over the confidentiality of sensitive data or the computing and storage constraints on the serving device. 
Some methods, such as GDE \cite{ICLR2022_GDE}, construct multiple source models using source samples in advance, thus eliminating the need for source access. 
However, training multiple models beforehand to estimate the accuracy of a deployed model is not practical as it is a common practice to train a single optimal model for deployment purposes.
To overcome these challenges, we propose a novel framework for estimating the accuracy of models on unlabeled target data without requiring access to source samples. 
Specifically, we derive a target labeling function that reduces the target risk from a pre-trained source model without using the source data. 
We then estimate the target accuracy of the models by computing the disagreement rate between the source and the target models under the target distribution.

Initially, we explore the viability of employing straightforward pseudo-labeling strategies \cite{PAWS,SwAV,FAUST,SHOT,FixMatch} that rely solely on target data. 
While a vanilla UDA method such as DANN requires access to both labeled source and unlabeled target data, recent advances in {\it source-free} UDA methods \cite{SFDA,FAUST,AAA,3C-GAN,SHOT,A2Net} remove the need for source data access. 
Certain source-free UDA methods \cite{FAUST, ISFDA, SHOT} freeze the head classifier of the source model and aim to train {\it target-specific} feature generators.
These feature generators are trained to learn target features that align with the source distribution of the frozen head classifier under the target distribution. 
These methods exhibit comparable performance to the state-of-the-art vanilla UDA and align with our goal of developing a {\it source-free UAE} approach based on pseudo-labeling.
To achieve this goal, we introduce the SF-DAP (Source-Free Domain-Adaptive Pseudo-labeling) framework, which incorporates source-free UDA algorithms into the source-free UAE.
As the adapted model by UDA algorithms may not always approximate an ideal target labeling function, computing disagreement in a naive manner can lead to a less accurate estimation.
To tackle this issue, we develop a systematic method that leverages perturbations, particularly virtual adversarial perturbations (VAPs) \cite{VAT}, to target data during inference.
Our domain-adaptive VAPs account for predictive uncertainty and domain discrepancy, thereby mitigating the effects of distribution shifts.
Extensive experimental results in various challenging scenarios demonstrate that our proposed method outperforms existing approaches without requiring source data.
Our key contributions are summarized as follows:
\begin{itemize}
\item
We propose SF-DAP, a source-free UAE framework that employs source-free UDA for a viable pseudo-labeling function under target distribution and combines target-adaptive VAPs with it. 
As far as we know, ours is the first source-free UAE approach that demonstrates comparable performance to source-based ones. 
\item
We illustrate the effectiveness of our proposed framework through extensive experiments on various challenging cross-domain scenarios.
Our approach consistently outperforms existing methods, even without labeled source data.
\item
We present Pseudo-labeling Assignment by Feature Alignment (PAFA) algorithm for source-free UDA, which extends existing methods by introducing modification that improves its efficacy.
Empirical results indicate that PAFA matches well with our framework.
\end{itemize}

\vspace{-2ex}

% ----------------------------------------------------------------------------------
\section{Preliminaries}
\subsection{Notation}
We use $\X$, $\Y$, and $\Z$ to denote the input, output, and feature representation space, respectively.
For simplicity of exposition, we consider binary {\it hypothesis} $h : \X \rightarrow \{0,1\}$ and $h$ is a composition of {\it feature generator} $g : \X \mapsto \Z $ and {\it classifier} $f : \Z \rightarrow \{0,1\}$ i.e., $h=f\circ g$. 
We define a {\it domain} as $\langle \D, h^*\rangle$. $\D$ denotes a distribution on input $\X$ and a labeling function $h^*$ is defined by $h^* : \X \rightarrow [0,1]$ ($h^*$ can have a fractional value).
The {\it error} of a hypothesis $h$ w.r.t. the labeling function $h^*$ under $\D$ is defined as $\varepsilon_{\D}(h, h^*) := \E_{\x \sim \D}\big{[}|h(\x)-h^*(\x)|\big{]} = \E_{\x \sim \D}\big{[}{\mathbbm 1}(h(\x)\neq h^*(\x))\big{]}$ and reduces to a disagreement probability such that $\varepsilon_{\D}(h, h^*) = \text{Pr}_{\x \sim \D}(h(\x)\neq h^*(\x))$.

$\varepsilon_{\D}(h)$ denotes the {\it risk} of $h$ that is the error of $h$ w.r.t. the true labeling function under $\D$, i.e., in case $h^*$ is the true labeling function, $\varepsilon_{\D}(h)=\varepsilon_{\D}(h,h^*)$.
To distinguish {\it source} and {\it target}, we attach subscripts S and T, respectively. {\eg}, $\D_S$ and $\D_T$ for the source and the target domain. 
For simplicity, we use $\varepsilon_S := \varepsilon_{\D_S}$ and $\varepsilon_T := \varepsilon_{\D_T}$.

% ---------------------------------------------------------------------------
\subsection{Unsupervised Accuracy Estimation}
One of the early UAE approaches \cite{CVPR2021_Deng} draws on the negative correlation between the distribution discrepancy and model accuracy and builds a regression model between these quantities for UAE. The correlation between the rotation estimation and the classification tasks is observed and used to build a simple regression-based UAE in the same manner, \cite{ICML2021_Deng}. 
AC and DoC \cite{ICCV2021_DoC} approach the problem from the prediction confidence. 
Similarly, ATC \cite{ICLR2022_ATC} utilizes confidence measures after calibration to identify the source data's confidence threshold that matches the source accuracy, which is applied to calculate the target accuracy\cite{Calibration_ICML17}. 
The initial UAE approach based on disagreement from random subsets \cite{Nakkiran_Bansal} was extended by GDE \cite{ICLR2022_GDE} to random model ensembles. 
With a flavor of the UDA approach, Proxy Risk \cite{ICML2020_DIR} explored the maximization of proxy risks, RI, and RM \cite{NEURIPS2021_RM} employed the iterative ensemble as well. 

\paragraph{Source-Free UAE.}
We consider the UAE problem where we have $m$ unlabeled target data $\{\x_j\}_{j=1}^m \in \X^m$ and a model (hypothesis) $h$ trained by $n$ labeled source samples $\{(\x_i,y_i)\}_{i=1}^n \in (\X \times \Y)^n$. 
In contrast to prior methods, we adopt a more practical and widely applicable assumption that the source samples are unavailable, as illustrated in Fig. \ref{fig:concept}.
The goal of {\it source-free} UAE is to find a function that correctly estimates accuracy, or equivalently, risk of the source model on unlabeled target data under $\D_T$ (denoted by $\varepsilon_T(h_S)$) without source samples.
We consider the case where both domains have the same set of labels.

% ---------------------------------------------------------------------------
\subsection{Unsupervised Domain Adaptation}
The goal of UDA is to find a hypothesis $h$ that correctly predicts the label $y_j$ of a new target sample $x_j$ by learning from the labeled source and unlabeled target data. 
The UDA problem considers that the learning algorithm has access to a set of $n$ labeled points $\{(\x_i,y_i)\}_{i=1}^n \in (\X \times \Y)^n \overset{\text{i.i.d.}}{\sim}\langle \D_S, h^*_S\rangle$ and a set of unlabeled points $\{\x_j\}_{j=1}^m \in \X^m \overset{\text{i.i.d.}}{\sim}\langle \D_T, h^*_T\rangle$.
Similarly to UAE, we assume both domains share the same label set.
DANN \cite{DANN} has received attention among earlier UDA works due to its approach to learning a domain-invariant representation with theoretical elegance.
DANN is also employed by some existing UAE methods.

\paragraph{Source-Free UDA.}
Source-free UDAs consider more practical limitations of source sample unavailability.
Unlike vanilla UDA, source-free UDA addresses the same problem without access to source samples, which aligns well with the nature of source-free UAE.
We pay close attention to an earlier work SHOT \cite{SHOT} which proposed a source hypothesis transfer strategy and achieved comparable performance to vanilla UDAs. 
ISFDA \cite{ISFDA} utilize SHOT to tackle class-imbalanced scenarios.
FAUST \cite{FAUST} also fixes the head classifier but produces improved performance by enforcing two consistency losses from multiple perturbed views of an input.
They typically learn a target-specific feature embedding $z=g(x)$ that can be aligned with the source distribution contained in the frozen (head) classifier $f$.

\subsection{Analysis on Existing DANN-Based Approaches}
\label{ssec:bounds}
We examine the theoretical framework associated with the recent approaches \cite{NEURIPS2021_RM, ICML2020_DIR} that learn domain-invariant representation using DANN \cite{DANN}.
Ben-David \etal \cite{Ben10} introduced ${\cal H}{\triangle}{\cal H}$-divergence 
which is determined by the discrepancy between source and target distributions of the hypothesis class ${\cal H}$. 
Given domains $\D_S$ and $\D_T$ over $\X$, the ${\cal H}\triangle{\cal H}$-divergence, $d_{{\cal H}\triangle{\cal H}}(\D_S,\D_T)$, is defined by $\sup_{h,h'\in{\cal H}} |\varepsilon_S (h, h') - \varepsilon_T (h,h')|$.
This divergence leads to 
\begin{theorem}[Ben-David \etal]
For any hypothesis $h\in{\cal H}$, let $\lambda = \min_{h'\in {\cal H}} \varepsilon_S(h') + \varepsilon_T(h')$ is the risk of ideal joint hypothesis. Then we have
\begin{equation}
\varepsilon_T(h) \leq \varepsilon_S(h) + d_{{\cal H}\triangle{\cal H}}(\D_S,\D_T) + \lambda.
\label{eq:bendavid}
\end{equation}
\end{theorem}
\noindent
Inequality \ref{eq:bendavid} implies that no classifier can accurately predict both domains when $\lambda$ is large, making it impossible to find a high-quality target hypothesis through UDA.
Later, Zhao \etal \cite{ICML2019_Zhao} presented an information-theoretic lower bound for the risk of ideal joint hypothesis as follows:
\begin{align}
\varepsilon_S+\varepsilon_T \geq & \frac{1}{2}(d_{JS}(\D_S^Y,\D_T^Y)-d_{JS}(\D_S^Z,\D_T^Z))^2,
\label{eq:zhao}
\end{align}
where $d_{JS}$, $\D^Y$, and $\D^Z$ denote Jensen-Shannon distance, label distribution, and feature distribution, respectively.

%------------------------------------------------------------------------
\begin{figure*}[!t]
  \centering
  \includegraphics[width=0.98\textwidth]{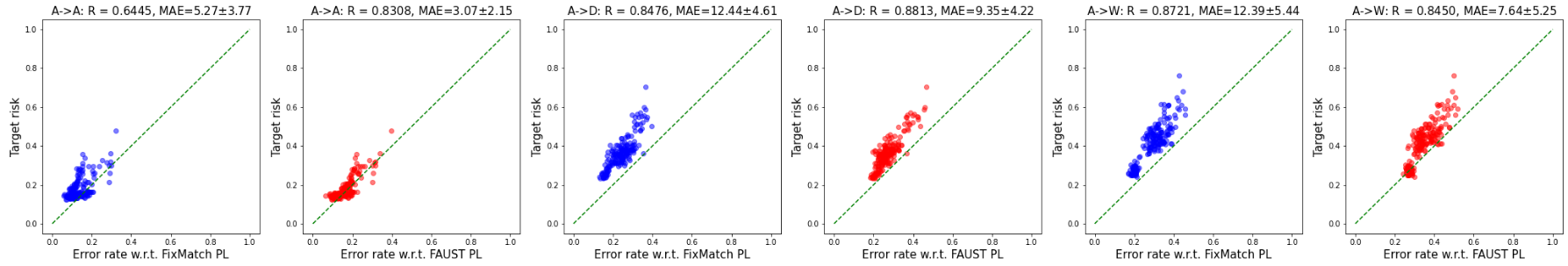}
%\vspace{-1.4ex}
\caption{
(Best viewed in color) Investigation on Office-31.
The scatter plots show the error rate w.r.t. pseudo-labels (x-axis) vs the true target risk (y-axis) of 192 different models trained on Amazon images. 
The green dashed line denotes the true target risk estimates. 
Each pair is tested on Amazon (test), DSLR, and Webcam, in that order.
Blue and red dots are produced based on \cite{FixMatch} and \cite{FAUST}, respectively.
}
\label{fig:PL_office}
\end{figure*}
%------------------------------------------------------------------------

By analyzing the inequality \ref{eq:bendavid} and the inequality \ref{eq:zhao}, we can identify three drawbacks of current approaches that use DANN.
Firstly, ${\cal H}\triangle{\cal H}$-divergence in the inequality \ref{eq:bendavid} cannot be accurately estimated from finite samples of arbitrary distributions, which limits its practical usefulness.
Secondly, as indicated in the inequality \ref{eq:zhao}, relying solely on learning domain-invariant representation can increase the lower bound of the risk of ideal joint hypothesis, which results in unsuccessful UDA and negatively impacts estimation performance. 
Thirdly, distance terms in the inequality \ref{eq:bendavid} and the inequality \ref{eq:zhao} require access to source samples, which does not align with the objective of a more practical source-free UAE.

% ---------------------------------------------------------------------------
\section{Straightforward Pseudo-Labeling Approach}
\label{sec:straightforward}

To achieve the goal of a source-free UAE, our focus is on determining how to bound $\varepsilon_T$ using terms that can be computed within the target domain.   
From the definition of risk and triangular inequality, for any hypothesis $h'_S\in{\cal H}$, the target risk $\varepsilon_T$ of the source hypothesis $h_S$ can be bounded as
$\varepsilon_T(h_S)\leq \varepsilon_T(h_S, h'_S)+\varepsilon_T(h'_S)$. 
Substituting $h'_S$ with a pseudo-labeling function $h^{pl}_T$ on $\D_T$ leads to:
\begin{proposition} 
Let $h^{pl}_T$ be any pseudo-labeling function under $\D_T$. Then, for given $h_S$, we have
\begin{equation}
\varepsilon_T(h_S) \leq \varepsilon_T(h_S, h^{pl}_T) + \varepsilon_T({h^{pl}_T}).
\label{eq:pseudo_label_ineq}
\end{equation}
\begin{proof}
By definition, $\varepsilon_T(h_S):=\E_{D_T}[|h_S-h^*_T|]$ where $h^*_T$ denotes true labeling function under $\D_T$.
Then, $\E_{D_T}[|h_S-h^*_T|]=\E_{D_T}[|h_S-h^{pl}_T+h^{pl}_T-h^*_T|]\leq\E_{D_T}[|h_S-h^{pl}_T|]+E_{D_T}[|h^{pl}_T-h^*_T|]$ from triangular inequality. 
Hence, $\varepsilon_T(h_S) \leq \varepsilon_T(h_S, h^{pl}_T) + \varepsilon_T({h^{pl}_T})$.
\end{proof}
\label{prop:pseudo_label_ineq}
\end{proposition}
\noindent
Proposition \ref{prop:pseudo_label_ineq} suggests that we can estimate the target accuracy of $h_S$ by identifying a suitable pseudo-labeling function for the target samples.  
This approach transforms our source-free UAE problem into the task of finding an effective pseudo-labeling function under $\D_T$ that reduces the estimation error.
It is worth noting that Proposition \ref{prop:pseudo_label_ineq} is different from the inequality \ref{eq:bendavid} or the inequality \ref{eq:zhao} in that the right-hand side can be computed only using the unlabeled target data in Proposition \ref{prop:pseudo_label_ineq}.
 
Based on Proposition \ref{prop:pseudo_label_ineq}, 
we investigate how well the disagreement between the source model's outputs $h_S$ and pseudo-labels on target data $h_T^{pl}$ matches true risk through experiments.
Among the previously proposed pseudo-labeling strategies \cite{PAWS,SwAV,FAUST,FixMatch}, 
we select (1) FixMatch \cite{FixMatch} approach that computes an artificial label by direct inference of the source model $h_S$ and (2) weighted feature prototype-based approach such as \cite{PAWS,SwAV,FAUST}. 
For the first approach, we obtain the model's predicted class distribution $q$ given a weakly perturbed view of an input, $x^w$, then use $h^{pl} \coloneqq \argmax(q(x^w))$ as a pseudo-label, which is compared with the model's output for a strongly perturbed view of the input, $q(x^s)$, as suggested by \cite{FixMatch}.
We also follow the approach of \cite{FAUST} that obtains the feature prototype of each class $k$ by $c^k \coloneqq \sum_{i\in \B} q^{k}(x_i) g(x_i)$ where $x_i$ is drawn from the empirical target data (in each mini-batch $\B$) and $q^{k}(x_i)$ is the prediction probability of $x_i$ to class $k$. 
Then, the soft pseudo-label of a target sample is defined by 
\begin{equation}
    h^{pl} \coloneqq \text{softmax}(C^T g(x))
    \label{eq:nfpc}
\end{equation}
where columns of the matrix $C$ are the prototypes.

\paragraph{Experiment.}
\label{ssec:pl_exp}
We investigate on Office-31 dataset which contains 4,652 ImageNet\cite{ImageNet}-like images across 31 objects collected from three different domains: Amazon (A), DSLR (D), and Webcam (W). 
We split the Amazon set into a development set (90\%) and a holdout set (10\%), then train 192 different ResNet50 source models  by sharing the development set but different training options such as learning rate, input augmentation, optimizer, and label smoothing. 
More details are described in Appendix \ref{app:exp_details}.
Webcam, DSLR, and Amazon holdout images are assigned as target data on which we evaluate the pseudo-label based estimators' performance.
The pseudo-labeling from weighted feature prototypes seems more noticeable as presented using red dots in Fig. \ref{fig:PL_office}. 
For the in-distribution setting (A$\rightarrow$A), the mean absolute error (MAE) of the estimator is 3.07$\pm$2.15. 
MAEs in the out-of-distribution settings (A$\rightarrow$D, A$\rightarrow$W) are no more than 9.35. 
Their Pearson's correlation coefficients are all above 0.83, which is close to the ideal correlation coefficient, 1.0.

%------------------------------------------------------------------------

\section{Proposed Method}
\label{ssec:SFUDA}

The observations presented in Sec \ref{sec:straightforward} corroborate our notion that the accuracy estimation performance can be enhanced by devising a pseudo-labeling function that more closely approximates the true labeling function under the target distribution. 
This concept logically leads to the application of source-free UDA to generate an appropriate pseudo-labeling function since source-free UDA aims to adapt the source model to enhance prediction accuracy on unlabeled target data without access to source data.

Let $h_T$ be a target model adapted to target distribution from the source model $h_S$ via a source-free UDA algorithm and therefore $h_T$ can be regarded as a pseudo-labeling function that has relatively less target risk. 
By replacing $h_T^{PL}$ with $h_T$, for any (adapted) target hypothesis $h_T\in{\cal H}$, Proposition \ref{prop:pseudo_label_ineq} comes to $\varepsilon_T(h_S) \leq \varepsilon_T(h_S, h_T) + \varepsilon_T(h_T)$.
This inequality indicates that the target risk of the source model $\varepsilon_T(h_S)$ becomes closer to the disagreement between source and target models under  $\D_T$ as the source-free UDA algorithm improves, i.e. the target risk of the adapted model $\varepsilon_T(h_T)$ becomes less.

However, UDA cannot fully approximate a true labeling function on $\D_T$ in general due to limitations imposed by the classification problem itself as acknowledged by \cite{ICML2020_DIR}. 
To tackle this limitation, we propose a target-adaptive pseudo-labeling technique that combines the pseudo-labeling function of each target sample with the inference over its $\epsilon$-neighbors that maximally disagrees with its own output. 
Then its disagreement with $h_S$ naturally estimates the desired target risk. Based on these ideas, we introduce a novel source-free UAE framework called {\bf SF-DAP} (Source-Free Domain-Adaptive Pseudo-labeling), which comprises source-free UDA and target-adaptive estimation. 

\subsection{Source-Free UDA Selection}
\label{ssec:selection}
For the selection of the source-free UDA algorithm, the empirical results in Sec \ref{sec:straightforward} allow us to consider two factors: 
(1) preservation of the ample information inherent in the classifier $f$ delivered from the source domain, which can confer an advantage in a source-free scenario, and 
(2) alignment of the target feature representation with the source distribution, as a proxy for successful adaptation, since pseudo-labeling in the feature space has demonstrated more favorable performance.

%------------------------------------------------------------------------
\begin{figure*}[!t]
  \centering
  \begin{minipage}[t]{0.20\textwidth} \centering \footnotesize
  \includegraphics[width=\linewidth]{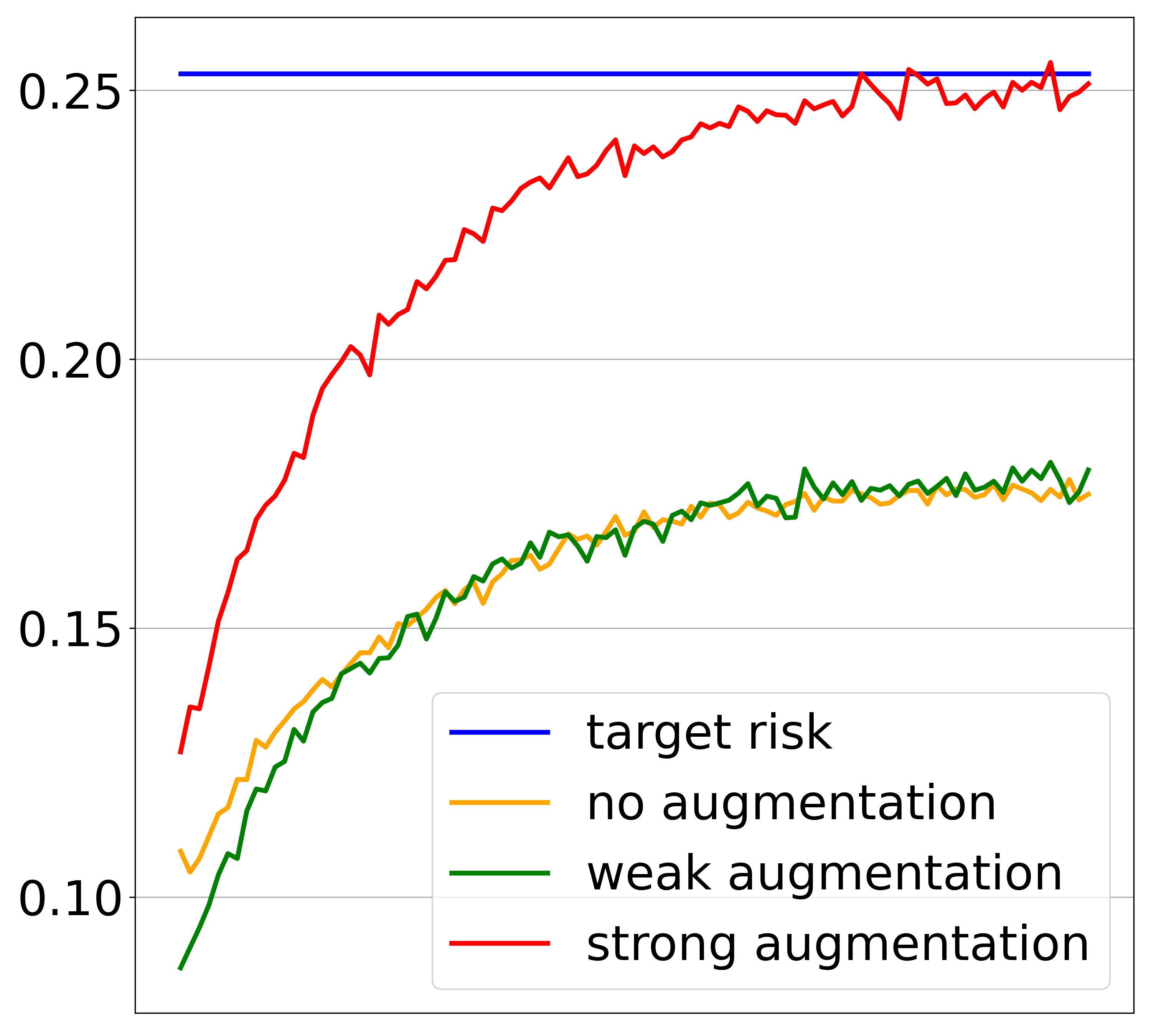}
  (a) Fog
  \end{minipage}~~
  \begin{minipage}[t]{0.20\textwidth} \centering \footnotesize
  \includegraphics[width=\linewidth]{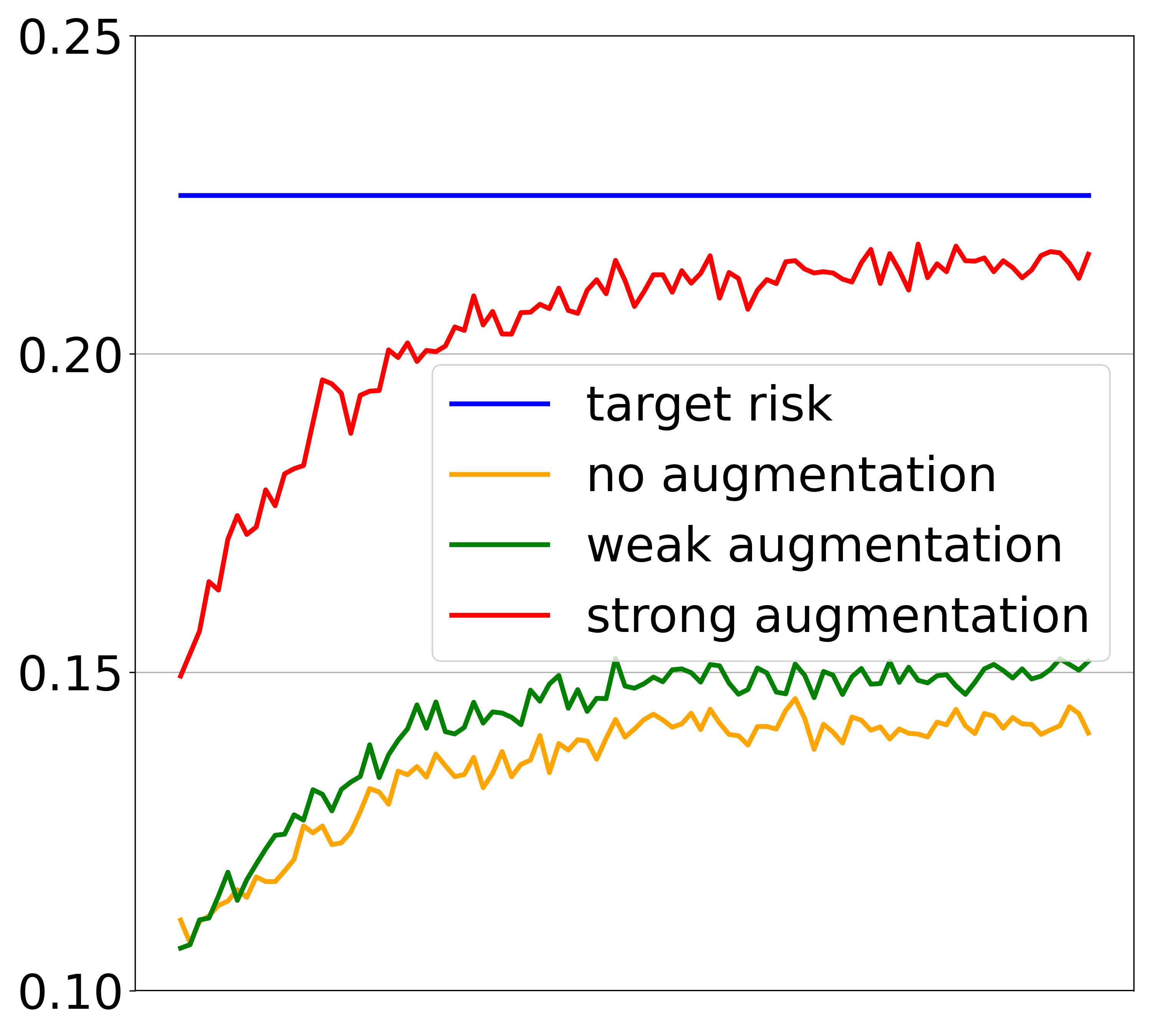}
  (b) Contrast 
  \end{minipage}~~
  \begin{minipage}[t]{0.20\textwidth} \centering \footnotesize
  \includegraphics[width=\linewidth]{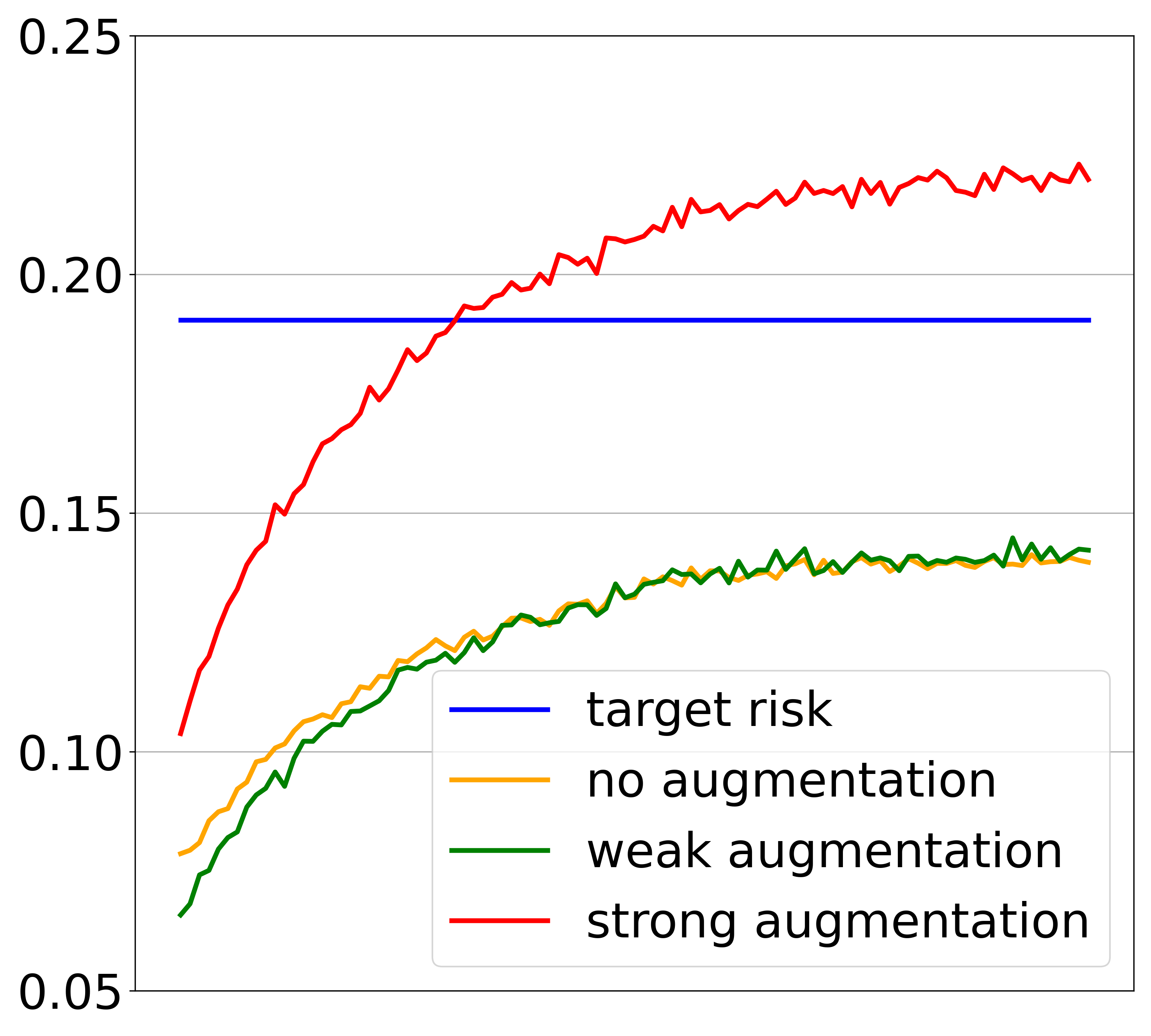}
  (c) Gaussian blur
  \end{minipage}~~
  \begin{minipage}[t]{0.20\textwidth} \centering \footnotesize
  \includegraphics[width=\linewidth]{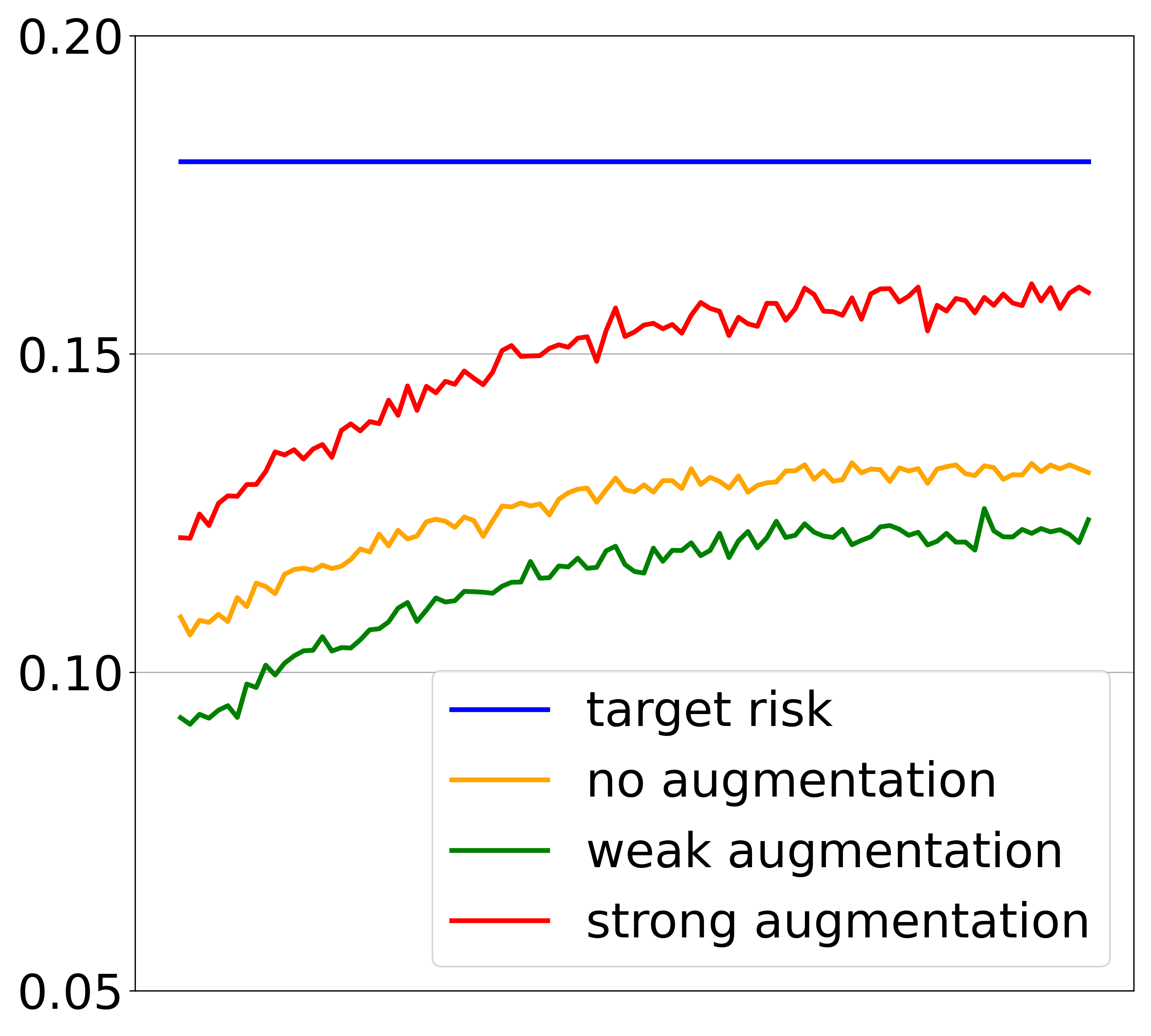}
  (d) Saturate 
  \end{minipage}
  %
%\vspace{-1.4ex}
\caption{
(Best viewed in color) 
The accuracy trends from different test-time perturbations as UDA progresses on CIFAR-10-C tasks.
The plot consists of the UDA training iteration ($x$-axis) vs. the target risk ($y$-axis). 
The blue line denotes the true target risk of $h_S$, and the other lines denote the estimated target risk of $h_S$ using the disagreement rate $\E_{x\sim\D_T}[|h_S(x)-h_T(x)|]$ where the input data $x$ are augmented in different fashions. 
In each red, green, and orange line, inputs are strongly augmented, weakly augmented, and not augmented, respectively.} 
\label{fig:aug_observation}
\end{figure*}
%------------------------------------------------------------------------

SHOT \cite{SHOT} freezes the head classifier $f$ and focuses on training the feature generator $g$, which satisfies the aforementioned criteria. 
Its simpler variant, SHOT-IM, trains the feature generator $g$ using an information maximization objective that combines entropy minimization loss and mean entropy maximization loss. 
However, SHOT requires augmentation of the source network prior to training (such as a bottleneck layer and two normalization layers), which does not align with our scenario of estimating the performance of an already trained source model. 
 Therefore, we introduce a modified version of SHOT in this work.
 
Instead of augmenting the network, we adopt a self-training approach that incorporates two input views into the training objective of SHOT-IM.
This approach is motivated by previous works \cite{PAWS, FAUST, FixMatch} and aims to alleviate the negative impact of not meeting network augmentation requirements. 
Specifically, we generate pseudo-labels based on the nearest feature prototype classifier defined in Eq. \ref{eq:nfpc} using the weakly perturbed view, while the strongly perturbed view is utilized for forward pass prediction.
We refer to the proposed source-free UDA training objective as PAFA (Pseudo-label Assignment by Feature Alignment) and formulate it as follows:
\begin{align}
    \min_{g}~ & {\cal H}(h(x^w)) + D_\text{KL}\left(\bar{h}(x^w) \big\| \frac{1}{K}{\bf 1}_K\right) -\log K \notag \\
    & + \alpha~ {\cal H}(h^{pl}(x^w), h(x^s)).  
    \label{eq:pafa}
\end{align}
Here, $x^w$ and $x^s$ denote weakly and strongly perturbed views of the input $x$, which are generated by standard flip-crop-rotation augmentation and by RandAugment \cite{RandAugment}, respectively.
The function ${\cal H}(\cdot)$ represents the Shannon entropy, while ${\cal H}(\cdot,\cdot)$ denotes cross-entropy. 
The term $\bar{h}(x)$ refers to the mean embedding of $K$-dimensional prediction outputs under the target distribution, and ${\bf 1}_K$ is a $K$-dimensional vector with all ones.
The positive constant $\alpha$ is a scaling factor for the self-training gradient. 
We note that the $\frac{1}{K}{\bf 1}_K$ term assumes that the target label distribution is uniform as SHOT \cite{SHOT} did. 
This assumption is reasonable since we lack any prior knowledge about the target label distribution.
Therefore, we also assume a uniform label distribution as a non-informative prior \cite{jeffreys1946invariant}. 
Nevertheless, our proposed method, PAFA,  performs well even with non-uniform target label distributions as demonstrated in Sec \ref{section:experiment}.
The UDA performance of PAFA is summarized in Sec \ref{sec:analysis}.
Additionally, we evaluate some other source-free UDA methods within the proposed framework and show there are no significant difference between their results  (see Appendix \ref{app:analysis}).

% ----------------------------------------------------------------------------------
\subsection{Enhancing the Quality of Estimation}
\label{ssec:enhancing}
An adapted model obtained through UDA, as presented in Sec \ref{ssec:selection}, can serve as a target labeling function in constructing the estimator using the following formulation:
\begin{equation}    
     \text{EST}_\text{naive} \coloneqq \E_{x\sim\D_T}[{\mathbbm 1}(h_S(x) \neq h_T(x))].
     \label{eq:est_naive}
\end{equation}  
Though the adapted model has reduced target risk when compared to the source model, it may not accurately approximate the true labeling function under a severe distribution shift.
Consequently, computing the disagreement between the source model $h_S$ and the adapted target model $h_T$ in a straightforward manner like Eq. \ref{eq:est_naive} often result in less accurate estimation.
Therefore, we need to explore methods to enhance the quality of estimation.

\subsubsection{Random Perturbation (RND)}
\label{sssec:random_perturb}
While evaluating the effectiveness of $\text{EST}_\text{naive}$ using CIFAR-10 \cite{CIFAR10} and CIFAR-10-C \cite{CIFAR10-C} benchmarks, we observed that applying a test-time perturbation to target samples before computing disagreement can improve the quality of estimation.
This observation is depicted in Fig. \ref{fig:aug_observation}, where we can see that the estimation quality often increases with the strength of the perturbation. 
Our finding suggests that perturbing the input to $h_T$ can reduce the estimation error, particularly when stronger perturbations (indicated by red lines in Fig. \ref{fig:aug_observation}) are applied, resulting in improved estimation. 
We can express this estimator as follows:
\begin{equation}    
     \text{EST}_\text{rnd} \coloneqq \E_{x\sim\D_T}[{\mathbbm 1}(h_S(x) \neq h_T(x^r))],  
     \label{eq:est_rand}
\end{equation}  
where $x^r$ denotes a randomly perturbed view of the input.
Though generating multiple perturbed views and ensembling inferences of them can further improve estimation performance (see Appendix \ref{app:analysis}), we have opted to present only a single perturbation for the sake of simplicity. 
As depicted in Fig. \ref{fig:aug_observation}c, random perturbations often lead to improved performance but may sometimes underestimates.

\subsubsection{Adversarial Perturbation (ADV)}
\label{sssec:adv_perturb}
Sec \ref{sssec:random_perturb} suggests that adding random perturbations to the inputs of the target model can frequently enhance the estimation performance, but this is not always the case. 
To ensure appropriate randomness during estimation, we consider the virtual adversarial perturbation (VAP) \cite{VAT}, which can adapt to the target distribution.
In addition to enforcing local smoothness of the conditional label distribution given unseen target inputs \cite{VAT}, VAP promotes the discovery of $\epsilon$-neighbors for each target sample that maximally disagrees with its own output. 
The disagreement between the pseudo-label function 
and the source model output naturally estimates the target risk. 
As an illustrative example: if a target sample resides far away from the decision boundaries, its pseudo-label will mostly agree with its own. 
However, if the sample is close to the decision boundaries, its adversarially estimated pseudo-label may indicate a different class label.
% } 
Hence, we define the accuracy estimator as follows:
\begin{equation}    
     \text{EST}_\text{adv} \coloneqq \E_{x\sim\D_T}\left[{\mathbbm 1}(h_S(x) \neq h_T(x+r_{vadv}))\right].  
     \label{eq:est_adv}
\end{equation}  
The virtual adversarial perturbation $r_{vadv}$ is computed as
\begin{equation}    
    r_{vadv} = \argmax_{\delta ~ s.t. ~ \|\delta\|_{2}\leq\epsilon} D_\text{KL}\Big{(}h_T(x)~\|~h_T(x+\delta)\Big{)},  
    \label{eq:est_adv_kld}
\end{equation}  
where $\epsilon$ is a hyperparameter that determines the magnitude of the adversarial perturbation.
We note that VAP is computed during inference through the frozen target model.
SF-DAP (ADV) comprises PAFA and $\text{EST}_\text{adv}$.

\subsubsection{Adaptive Adversarial Perturbation (AAP)}
\label{sssec:aadv_perturb}
By employing Eq. \ref{eq:est_adv}, we achieved estimation performance comparable to that of existing methods that require access to source samples, as presented in Sec \ref{section:experiment}. 
Nonetheless, we identified certain concerns, including instances where the estimation errors were unacceptably large in specific scenarios. 
The proposed method is able to address this issue by adjusting the sole hyperparameter $\epsilon$.
This section outlines how to quantify the factors influencing the magnitude of $\epsilon$ and concludes with an improved formulation.

%------------------------------------------------------------------------
\begin{table*}[!t]
\centering
\def\arraystretch{1.3}
\caption{ 
Overall benchmark results on various UAE tasks. 
Average mean absolute errors (MAEs, \%) of each group and overall averages are reported. 
{\bf Bold} numbers indicate the superior results and {\bf \textit{bold-italic}} do the next best results. Full results are presented in Appendix \ref{app:exp_details}.
}
%\vspace{-1.4ex}
\resizebox{0.96\textwidth}{!}
{%
\begin{tabular}{c| c| c|c c c c c|c c}
\Xhline{2\arrayrulewidth}
\multicolumn{3}{c|}{}  & \multicolumn{5}{c|}{source access approach} & \multicolumn{2}{c}{source-free approach (ours)}  \\
\hline
datasets & settings & subset description
    & DoC \cite{ICCV2021_DoC}    & Proxy \cite{ICML2020_DIR}     & RI \cite{NEURIPS2021_RM} &RM \cite{NEURIPS2021_RM} & GDE \cite{ICLR2022_GDE}   
    & {SF-DAP (ADV)} & {SF-DAP (AAP)}\\

\hline
Digits & 6 & MNIST, USPS, SVHN
    & 14.62$\pm$0.80	&9.63$\pm$1.44	&9.50$\pm$1.51	&11.33$\pm$1.36	&27.80$\pm$0.75	&{\bf \textit{2.18$\pm$0.56}}	&{\bf 2.15$\pm$0.58} \\
Office-31 & 6 & Amazon, DSLR, Webcam
    & 5.00$\pm$1.26	    &4.64$\pm$1.23	&8.69$\pm$1.62	&{\bf \textit{2.73$\pm$1.28}}	&9.79$\pm$1.10	&4.72$\pm$1.13	&{\bf {2.51$\pm$1.10}} \\
Office-Home & 12 & Art, Clipart, Product, Real-World
    & 25.62$\pm$0.50	&8.75$\pm$1.41	&12.13$\pm$1.58	&{\bf 3.66$\pm$1.25}	&37.52$\pm$0.51	&8.73$\pm$0.65	&{\bf \textit{4.98$\pm$0.68}} \\
VisDA & 1 & Synthetic images, MSCOCO
    & 15.72$\pm$4.35	&8.90$\pm$1.71	&7.50$\pm$4.70	&{\bf \textit{4.41$\pm$2.52}}	&29.31$\pm$3.73	&{\bf \textit{4.41$\pm$1.10}}	&{\bf 1.73$\pm$0.93} \\
\hline
CIFAR-10 & 19 & CIFAR-10 and CIFAR-10-C subsets
    & 18.91$\pm$1.56	&10.47$\pm$0.83	&3.20$\pm$0.94	&{\bf 2.17$\pm$1.18}	&2.99$\pm$1.24	&{\bf \textit{2.34$\pm$1.47}}	&{ 3.06$\pm$1.46} \\
CIFAR-100 & 19 & CIFAR-100 and CIFAR-100-C subsets
    & 48.64$\pm$1.99	&26.34$\pm$2.46	&5.84$\pm$1.67	&{\bf 1.90$\pm$0.84}	&6.33$\pm$1.75	&6.81$\pm$1.59	&{\bf \textit{3.28$\pm$1.35}} \\
\hline
\multicolumn{3}{c|}{micro average (of all 63 settings) } 
    &27.37$\pm$1.50 &14.27$\pm$1.67 &6.89$\pm$1.46 &{\bf 3.33$\pm$1.14} &14.00$\pm$1.29 &5.15$\pm$1.29 &{\bf {3.33$\pm$1.20}}\\
\hline
\multicolumn{3}{c|}{macro average (of the above 6 averages) } 
    &21.42$\pm$1.32 &11.45$\pm$1.23 &7.81$\pm$1.42 &{\bf \textit{4.37$\pm$1.18}} &18.96$\pm$1.23 &4.86$\pm$1.04 &{\bf 2.95$\pm$1.01}\\
\Xhline{2\arrayrulewidth}
\end{tabular}
}

\vspace{-1ex}

\label{tab:main_result} 
\end{table*}
%------------------------------------------------------------------------

\paragraph{Predictive Uncertainty.}
According to the cluster assumption \cite{cluster_assumption}, data points with low predictive uncertainty are typically located far from the class decision boundary, resulting in more accurate model predictions. 
Conversely, data points with high predictive uncertainty often reside in the low-density regions of the feature space, near or overlapping the class decision boundary, leading to less accurate model predictions.
Therefore, to enhance estimation performance, we suggest reducing perturbation in data points with low uncertainty while increasing perturbation in those with high uncertainty.
This can be achieved by scaling the perturbation magnitude based on the predictive uncertainty measurement. 
We define the uncertainty factor $C_{unc}$ as the standard deviation (std) of the class probabilities predicted by the target model that ranges within 0.5.
To evaluate uncertainty, we apply Monte-Carlo dropout inference sampling $n$-times (see Appendix \ref{app:analysis} for more details):
\begin{equation}    
    C_{unc} = \text{std}\left\{q_i^{(k)}\right\}_{i=1}^n,
\end{equation}  
where $q_i^{(k)}$ is the probability of the $k$-th class in the predicted class distribution $q$ obtained from the $i$-th Monte-Carlo dropout sample.
The class index $k$ is determined by $\argmax_{j\in\{1,\cdots,K\}} \left(\frac{1}{n}\sum_{i=1}^n q_i\right)^{(j)}$.

\paragraph{Domain Divergence}
Though the $C_{unc}$ incorporates the cluster assumption into individual data points, bias may exist due to inaccurate predictions stemming from the distribution gap between the source and the target. 
To tackle this bias, we consider the divergence between both distributions.
We cannot access the source dataset in the source-free scenarios, so we  compute the divergence from the individual data point perspective instead of focusing on datasets. 
Hence, we define the divergence factor $C_{div}$ using Jensen-Shannon Divergence ($D_\text{JS}$) between $h_S(x_t)$ and $h_T(x_t)$:  
\begin{equation}    
    C_{div} = D_\text{JS}(h_S(x_t)\|h_T(x_t)).
\end{equation}  
By selecting base-2 logarithm, the $D_\text{JS}$ value is bounded between 0 and 1, being able to provide normalized adjustment to alleviate bias.
Adding this factor to $C_{unc}$ yields an {\it adjusted predictive uncertainty} factor $C_{adj\_unc}$.

\paragraph{Data Volume Density}
The density of data points in a dataset is often a crucial factor in determining the magnitude of the VAP. 
For instance, images that are synthetically contrast reduced have an increased color space density, while natural images typically have a much lower density than synthetic images.
We can enhance the estimation performance by adjusting the magnitude of VAP in proportion to the relative density of the data points. 
To this end, we consider the standard deviations calculated for each axis of the three-dimensional RGB color space that can be easily computed from the image dataset.
For the inaccessible source dataset, these values are still commonly available in the definition of data transform used for image normalization during training.
The natural adjustment approach scales the perturbation magnitude by the relative density of the target data to that of the source data.
For consistent application to both black-and-white and color images, we define the density scaling factor $C_{den}$ as follows: 
\begin{equation}    
    C_{den} = \frac{1}{3}\left(\frac{std_t^R}{std_s^R} + \frac{std_t^G}{std_s^G} + \frac{std_t^B}{std_s^B}\right),
\end{equation}  
where $std_s^R$ and $std_t^G$ denote the standard deviation of the source dataset in the red axis and that of the target dataset in the green axis, respectively ($B$ denotes blue).

\paragraph{Class Complexity}
We also consider the number of classes ($K$) that affects the scale of the VAP. KL divergence tends to increase as $K$ increases, as the volume of the probability space grows exponentially with $K$. While the behavior depends on specific distributions, our empirical analysis has shown that compensating for the increase of the KL divergence in Eq. \ref{eq:est_adv} by log$K$ improves the accuracy. 
Based on this observation, we introduce the class complexity factor $C_{cls}$:
\begin{equation}    
    C_{cls} = \log \left(\text{the number of categories}\right).
\end{equation}

\paragraph{Proposed Accuracy Estimator.}
Considering the factors mentioned above, we propose $\text{EST}_\text{aap}$ that employs adaptive adversarial perturbations (AAPs).
$\text{EST}_\text{aap}$ is, in fact, a modification of
$\text{EST}_\text{adv}$ in Eq. \ref{eq:est_adv}, with the only difference being the definition of hyperparameter $\epsilon$ as: 
\begin{equation}    
     \epsilon \coloneqq \epsilon_0 C_{cls} C_{den} C_{adj\_unc}.  
     \label{eq:est_aap}
\end{equation}  
Here, the base constant $\epsilon_0$ is typically set to 1.0 as suggested by \cite{VAT}.
According to Eq. \ref{eq:est_aap}, $\epsilon$ is re-scaled by the adjusted predictive uncertainty ($C_{adj\_unc}$) that adaptively controls the perturbation on individual data points. 
Then the updated $\epsilon$ is multiplied by the class complexity ($C_{cls}$) of the target dataset and by the relative data point density of the target dataset compared to the source dataset ($C_{den}$).
Thus, PAFA and $\text{EST}_\text{aap}$ collaboratively compose SF-DAP (AAP).

%------------------------------------------------------------------------
% \begin{table}
\begin{table*}[!t]
\centering
\def\arraystretch{1.1}
\caption{ 
Ablation study on Office-31. 
Uniformly scaling $\epsilon$ for all data instances within the dataset can hurt performance (Configuration 0).
With instance-wise scaling ($C_{adj\_unc}$), incorporation of scaling factors gradually improves accuracy estimation overall.
}
%\vspace{-1.4ex}
\resizebox{0.86\textwidth}{!}
{%
\begin{tabular}{c | l| c c c c c c | c }
\Xhline{2\arrayrulewidth}
Method& VAP magnitude ($\epsilon$) 
    & A$\rightarrow$D & A$\rightarrow$W & D$\rightarrow$A 
    & D$\rightarrow$W & W$\rightarrow$A & W$\rightarrow$D & Avg.\\
\hline
SF-DAP (ADV)& $\epsilon_0$(=1.0)&
1.85$\pm$1.22&	4.43$\pm$2.10&	9.05$\pm$1.78&	2.97$\pm$0.52&	9.56$\pm$1.77&	0.46$\pm$0.24& 4.72$\pm$1.13\\
Configuration 0 &$\epsilon_0 C_{den} C_{cls}$&
9.39$\pm$4.01&	5.64$\pm$5.20&	5.24$\pm$3.02&	7.64$\pm$2.75 &	{\bf 2.75$\pm$2.53} &	3.90$\pm$1.47 & 5.76$\pm$1.78\\
Configuration 1 &$\epsilon_0 C_{adj\_unc}$&
3.33$\pm$1.12&	1.85$\pm$1.14&	10.47$\pm$1.92&	{\bf 0.34$\pm$0.28}&	11.98$\pm$1.80&	0.22$\pm$0.14& 4.70$\pm$1.03\\
Configuration 2 &$\epsilon_0 C_{den} C_{adj\_unc}$&
4.70$\pm$1.13&	2.15$\pm$1.39&	8.20$\pm$1.93&	0.57$\pm$0.27&	10.82$\pm$1.58&	{\bf 0.16$\pm$0.15}& 4.43$\pm$1.04\\
Configuration 3 &$\epsilon_0 C_{cls} C_{adj\_unc}$&
1.53$\pm$0.76&	2.00$\pm$1.09&	4.84$\pm$1.82&	2.63$\pm$0.71&	6.03$\pm$1.99&	1.25$\pm$0.56& 3.05$\pm$1.07\\
SF-DAP (AAP)&$\epsilon_0 C_{cls} C_{den} C_{adj\_unc}$&
{\bf 0.96$\pm$0.87}&	{\bf 1.62$\pm$0.98}&	{\bf 3.28$\pm$1.97}&	3.08$\pm$0.89&	5.23$\pm$2.03&	0.88$\pm$0.49& {\bf 2.51$\pm$1.10}\\

\Xhline{2\arrayrulewidth}
\end{tabular}
}%
\vspace{-1ex}
\label{tab:ablation}
\end{table*}
%------------------------------------------------------------------------

\section{Experiment}\label{section:experiment}

\paragraph{Setup.}
We evaluate and report absolute estimation error to measure the performance of UAE.
For natural distribution shift, we experiment on Digits \cite{USPS,MNIST,SVHN}, Office-31 \cite{Office31}, Office-Home \cite{Office-Home}, and VisDA \cite{Visda17} datasets referring to the UDA literature.
We note that a few challenging scenarios are included, such as MNIST$\rightarrow$SVHN and VisDA sim-to-real.
For synthetic distribution shift setting, we use CIFAR-10 and CIFAR-100 \cite{CIFAR10}, paired with CIFAR-10-C and CIFAR-100-C\cite{CIFAR10-C}, respectively.
The CIFAR-10-C and CIFAR-100-C datasets contain 19 distinct types of corruptions applied to CIFAR-10 and CIFAR-100 images, respectively.
We employed an identical backbone for each scenario across all methods for a fair comparison. 
To be specific, we use a LeNet variant for Digits, ResNet18 for CIFAR-10 and CIFAR-100, ResNet50 for Office-31 and Office-Home, and ResNet101 for VisDA.
More detailed experimental setups, including datasets, network architectures, and training options, are described in Appendix \ref{app:exp_details}.

%------------------------------------------------------------------------
\begin{figure}[!t]
  \centering
  \includegraphics[width=0.95\columnwidth]{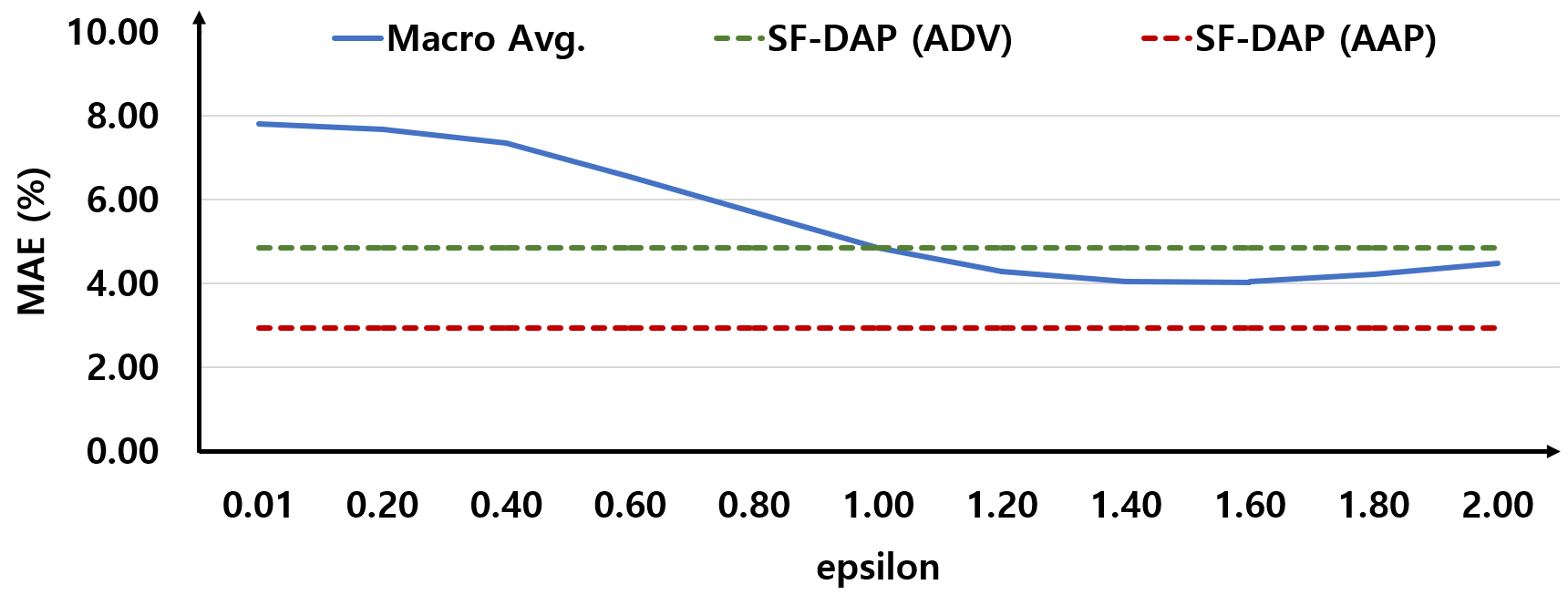}
\vspace{-1ex}
\caption{
(Best viewed in color) SF-DAP (AAP) outperforms the optimal result (at $\epsilon\approx 1.5$) achieved by uniform $\epsilon$ scaling.
}
\label{fig:MAE_trend_on_eps}
\end{figure}
%------------------------------------------------------------------------

\paragraph{Evaluation.}
As our baselines for comparison, we consider Difference of Confidence (DoC) \cite{ICCV2021_DoC}, Proxy Risk (Proxy) \cite{ICML2020_DIR}, Random Initialization (RI) \cite{NEURIPS2021_RM}, Representation Matching (RM) \cite{NEURIPS2021_RM}, and Generalization Disagreement Equality (GDE) \cite{ICLR2022_GDE}.
We re-implement all baselines and compare them under the same experimental setup.
We repeat experiments on ten independently-trained source models and report the mean and standard deviation of the computed MAE.
Evaluation protocols of the baselines and their details are described in Appendix  \ref{app:exp_details} and Appendix \ref{app:methods}, respectively.

\paragraph{Results.}
As shown in Table \ref{tab:main_result}, our proposed framework outperforms existing methods in 63 cross-domain scenarios across six benchmark groups, even without the need to access source samples.
The proposed SF-DAP (AAP) demonstrates comparable or superior results to the state-of-the-art methods like RM and RI. 
Despite its simpler approach, SF-DAP (ADV) also exhibits comparable or often better performance than previous approaches that require source sample access.
While some existing methods such as GDE \cite{ICLR2022_GDE}, RI, and RM \cite{NEURIPS2021_RM} show strong results on synthetic distribution shift scenarios, they have limitations under natural distribution shift settings.
We emphasize that our proposed framework requires only unlabeled target data,
whereas all these baseline methods cannot be applied if additional access to source samples is prohibited. 
All detailed results of 63 scenarios are presented in Appendix \ref{app:exp_details}.

\section{Analysis}
\label{sec:analysis}
\paragraph{Ablation.}
We conduct an ablation study on the Office-31 dataset by incrementally introducing scaling factors of SF-DAP (AAP), starting from SF-DAP (ADV).
Configuration 0 in Table \ref{tab:ablation} indicates that including the adjusted uncertainty factor ($C_{adj\_unc}$) is critical without prior knowledge of $\epsilon_0$.
Given $C_{adj\_unc}$, both the class complexity ($C_{cls}$) and the data volume density ($C_{den}$) factors contribute substantially to enhancing the estimation performance.
As a consequence, SF-DAP (AAP) surpass other configurations.
Additional ablation studies can be found in Appendix \ref{app:analysis}.

\paragraph{MAE on a Smooth $\epsilon$-Axis.}
We investigate the MAE trajectory over a smooth $\epsilon$-axis as depicted in Fig. \ref{fig:MAE_trend_on_eps}, with the blue line representing the macro average MAE across all 63 scenarios.
Fig. \ref{fig:MAE_trend_on_eps} suggests that an optimal estimation can be realized at $\epsilon \approx 1.5$ under a uniform VAP scaling constraint applied to all target datasets.
Though SF-DAP (ADV) employing $\epsilon = 1.0$ (green dashed line) might be sub-optimal, SF-DAP (AAP) -- illustrated by the red dashed line -- further diminishes MAE beyond the optimal MAE at $\epsilon\approx 1.5$.
This reduction is achieved by introducing $C_{adj\_unc}$ which enables adaptive VAP scaling per target instance, contrasting with the fixed scaling of $C_{cls}$ and $C_{den}$.

%------------------------------------------------------------------------
% \begin{table}
\begin{table}[!t]
\centering
\def\arraystretch{1.1}
\caption{ 
Comparison of PAFA with SHOT-IM \cite{SHOT} and FAUST \cite{FAUST} on the six UDA benchmark tasks.
{\it Source-only} denotes the target accuracy of the source model without applying UDA.
(S: SVHN, M: MNIST, U:USPS, Syn: SYNSIG, G:GTSRB)
}
%\vspace{-1.4ex}
\resizebox{1.0\columnwidth}{!}
{%
\begin{tabular}{c | c c c c c c | c}
\Xhline{2\arrayrulewidth}
Method 
    & S$\rightarrow$M & M$\rightarrow$S & M$\rightarrow$U 
    & U$\rightarrow$M & Syn$\rightarrow$G & VisDA & Avg.\\
% \midrule
\hline
Source only
    & 70.0   & 47.8   & 77.4    & 84.9   & 77.1   & 56.6  & 69.0\\
PAFA     
    % & {\bf 99.5$\pm$0.1}   & 83.5$\pm$0.3   & 97.8$\pm$0.0   
    % & {\bf 98.5$\pm$0.1}   & 99.4$\pm$0.0   & 83.8$\pm$0.3  & 93.8   \\
    & {\bf 99.5}   & 83.5   & 97.8   
    & {\bf 98.5}   & 99.4   & 83.8  & 93.8   \\
SHOT-IM      
    % & 98.5$\pm$0.8   & 11.0$\pm$0.5   & 97.8$\pm$0.4   
    % & 97.6$\pm$0.1   & 97.2$\pm$0.1   & 82.0$\pm$0.5    & 80.7 \\
    & 98.5   & 11.0   & 97.8   
    & 97.6   & 97.2   & 82.0    & 80.7 \\
FAUST
    % & 99.2$\pm$0.2   & {\bf 88.0$\pm$0.6}   & {\bf 98.9$\pm$0.1}   
    % & 97.6$\pm$0.1   & {\bf 99.7$\pm$0.0}   & {\bf 84.0$\pm$0.5}    & 94.6   \\
    & 99.2   & {\bf 88.0}   & {\bf 98.9}   
    & 97.6   & {\bf 99.7}   & {\bf 84.0}    & 94.6   \\
\Xhline{2\arrayrulewidth}
\end{tabular}
}%
\vspace{-1ex}
\label{tab:pafa}
\end{table}
%------------------------------------------------------------------------

\paragraph{Applying AAP Directly to the Source Model.}
In the application of AAP to the source model (without UDA) for Digits and Office-31 datasets, a notable 3.0\% increase is observed in average MAE compared to our SF-DAP (AAP) that incorporates UDA. 
This result underlines the crucial role of the source-free UDA element within the proposed SF-DAP framework. 

\paragraph{Estimation Time.}
We compare the runtime of various methods that require additional training.
Our method shows a competitive running time to other methods. 
In particular, the proposed SF-DAP framework runs faster than RM and Proxy Risk, which also perform UDA during estimation (see Appendix \ref{app:analysis} for details).

\paragraph{PAFA Performance in UDA.}
To contribute to the proposed SF-DAP framework as a source-free UDA algorithm, it is important for PAFA to produce a comparable performance to the existing state-of-the-art methods like SHOT-IM \cite{SHOT} and FAUST \cite{FAUST}.
We evaluated PAFA's performance on six commonly used UDA benchmarks and found that it achieved comparable results to SHOT-IM and FAUST as shown in Table \ref{tab:pafa}.
Notably, when adapting from an MNIST (source) model to the SVHN (target) dataset, PAFA achieved a target risk of less than 20\%, which is the state-of-the-art result in UDA literature.

\paragraph{UAE Performance during UDA.}
We track the performance trend of accuracy estimation as the proposed source-free UDA, PAFA, progresses to each target domain.
Our proposed framework, SF-DAP, starts producing accurate estimates surprisingly early and remains steady throughout the rest of the UDA iterations as illustrated in Fig \ref{fig:officehome_trend} and Fig \ref{fig:others_trend}.

% ----------------------------------------------------------------------------------
\section{Conclusion}
In this work, we proposed a novel framework that estimates model accuracy under unseen target distributions without access to source data. 
Our approach utilized source-free UDA algorithms to produce a viable pseudo-labeling function and employ virtual adversarial perturbations to enhance the accuracy estimation. 
We also developed a simple yet effective source-free UDA algorithm, PAFA, which achieved comparable performance to existing methods on six popular UDA benchmarks,
while avoiding the need for network augmentation of the source model.
To further improve the estimation quality, we introduced adaptive adversarial perturbations based on adjusted predictive uncertainty and domain discrepancy information. 
Our experimental evaluation demonstrated that both AAP and ADV configurations in our proposed framework outperformed existing methods.
To the best of our knowledge, this is the first study to address source-free scenarios in the UAE literature.

{\small
\bibliographystyle{ieee_fullname}
\bibliography{egbib}
}

%%%%%%%%%%%%%%%%%%%%%%%%%%%%%%%%%%%%%%%%%%%%%%%%%%%%%%%%%%%%

% \newpage
\appendix
\section{Experimental Details}
\label{app:exp_details}
\subsection{Datasets}
We evaluate the effectiveness of our proposed framework on several benchmark datasets including Digits, Office-31, Office-Home, and VisDA for natural distribution shift scenarios.
Additionally, we conduct experiments on synthetic distribution shift scenarios by employing CIFAR-10-C and CIFAR-100-C datasets.

\paragraph{Digits.}
We conducted experiments on smaller Digits datasets: MNIST \cite{MNIST}, USPS \cite{USPS}, and SVHN \cite{SVHN}.
These datasets comprise labeled images of digits ranging from 0 to 9. 
MNIST dataset comprises 60K training images and 10K test images, while USPS dataset contains 7,291 training images and 2,007 test images. 
SVHN dataset comprises 73,257 training images and 26,032 test images. 
All images are resized to 32$\times$32 during training and testing.
We considered all six cross-domain tasks, including the challenging MNIST$\rightarrow$SVHN and MNIST$\rightarrow$USPS scenarios. 

\paragraph{Office-31.}
We also evaluate Office-31 \cite{Office31} datasets.
The Office-31 dataset contains 4,652 images, similar to ImageNet \cite{ImageNet}, distributed across 31 objects from three distinct domains, namely, Amazon (A), DSLR (D), and Webcam (W), with 2,817, 498, and 795 images, respectively. 
For training and testing, we resized each domain set to 224$\times$224 after splitting it into a development set (90\%) and a holdout set (10\%). 
We evaluated all six source-target combinations.

\paragraph{Office-Home.}
The Office-Home \cite{Office-Home} dataset consists of 15,500 high-resolution images belonging to 65 categories from four distinct domains: Artistic images (Ar), Clip Art (Cl), Product images (Pr), and Real-World images (Rw). 
In this study, we considered all possible 12 UDA tasks.

\paragraph{VisDA.}
The VisDA \cite{Visda17} dataset contains a large-scale collection of complex images from 12 categories that include 152K synthetically rendered images from different angles and lighting configurations and 55K real-world images that are sampled from MSCOCO \cite{MSCOCO}. 
These images are resized to 224$\times$224 for training and testing.
Due to its more practical sim-to-real UDA problem, this dataset is suitable for more challenging cross-domain benchmarks.

\paragraph{CIFAR-10/CIFAR-10-C and CIFAR-100/CIFAR-100-C.}
CIFAR-10 and CIFAR-100 datasets \cite{CIFAR10} consist of 50K training images and 10K test images with 32$\times$32 dimensions per image, and ten and hundred mutually exclusive classes, respectively.
CIFAR-10-C and CIFAR-100-C datasets \cite{CIFAR10-C} introduce 19 types of artificial corruptions such as blur, noise, weather, and digital categories, creating corresponding corrupted subsets.
These CIFAR-10 to CIFAR-10-C and CIFAR-100 to CIFAR-100-C settings are quite suitable for measuring the UAE performance against diverse synthetic distribution shift scenarios.

\subsection{Network Architectures}
For Digits experiments, we used a LeNet variant for Digits with a convolutional kernel size of 5$\times$5 and applied ReLU function after each Batch Normalization (BN) layer. 
The entire architecture consists of:
Conv(3, 64) - BN(64) - Maxpool(2) - Conv(64, 64) - BN(64) - Maxpool(2) - Conv(64, 128) - BN(128) - Linear(8192, 3072) - BN(3072) - Linear(3072, 2048) - BN(2048) - Linear(2048, 10).

For CIFAR-10 / CIFAR-10-C and CIFAR-100 / CIFAR-100-C datasets, we employed the ResNet18 backbone, while we used the ResNet50 backbone for Office-31 and Office-Home datasets. 
The ResNet101 backbone was applied for VisDA.
The complete network comprises each backbone (i.e. from the initial layer to the global average pooling layer) and two additional fully connected layers.

\subsection{Baselines}
We consider Proxy Risk (Proxy) \cite{ICML2020_DIR}, Difference of Confidence (DoC) \cite{ICCV2021_DoC}, Random Initialization (RI) \cite{NEURIPS2021_RM}, Representation Matching (RM) \cite{NEURIPS2021_RM}, and Generalization Disagreement Equality (GDE) \cite{ICLR2022_GDE} as our baselines.
We re-implement all baselines and conducted a comparative analysis under identical experimental conditions.
It is worth noting that while these baselines necessitate access to source samples, our framework only requires unlabeled target samples when estimating the target risk of the source model.

%------------------------------------------------------------------------
\begin{table*}[!t]
\centering
\def\arraystretch{1.3}
\caption{ 
Full UAE benchmark results on natural distribution shift scenarios (MAE, \%). 
{\bf Bold} for the best and {\bf \textit{bold-italic}} for the next best. 
}
%\vspace{-1.4ex}
\resizebox{0.96\textwidth}{!}
{%
\begin{tabular}{c| c| c|c c c c c|c c}
\Xhline{2\arrayrulewidth}
\multicolumn{3}{c|}{}  & \multicolumn{5}{c|}{source access approach} & \multicolumn{2}{c}{source-free approach (ours)}  \\
\hline
datasets & source & target
    & DoC \cite{ICCV2021_DoC}    & Proxy \cite{ICML2020_DIR}     & RI \cite{NEURIPS2021_RM} &RM \cite{NEURIPS2021_RM} & GDE \cite{ICLR2022_GDE}   
    & {SF-DAP (ADV)} & {SF-DAP (AAP)}\\

\hline
\multirow{6}*{Digits} & MNIST & USPS
    &0.63$\pm$0.02	&0.45$\pm$0.34	&3.87$\pm$2.45	&0.55$\pm$0.25	&1.45$\pm$0.02	&1.17$\pm$0.05	&{\bf 0.05$\pm$0.04}\\
 & MNIST & SVHN
    &29.14$\pm$0.21	&20.08$\pm$2.64	&7.31$\pm$1.29	&28.07$\pm$1.70	&58.46$\pm$0.17	&{\bf 3.02$\pm$0.26}	&5.39$\pm$0.21\\
 & USPS & MNIST
    &12.83$\pm$0.12	&6.70$\pm$2.87	&29.10$\pm$2.60	&10.16$\pm$4.03	&33.65$\pm$0.10	&{\bf 0.40$\pm$0.14}	&1.25$\pm$0.29\\
 & USPS & SVHN
    &32.80$\pm$0.17	&18.97$\pm$2.96	&9.59$\pm$2.04	&21.97$\pm$3.01	&65.20$\pm$0.10	&6.16$\pm$0.21	&{\bf 3.84$\pm$0.28}\\
 & SVHN & MNIST
    &4.08$\pm$2.21	&6.22$\pm$1.93	&5.20$\pm$3.72	&3.28$\pm$1.29	&2.77$\pm$1.95	&1.38$\pm$0.50	&{\bf 0.99$\pm$0.54}\\
 & SVHN & USPS
    &8.25$\pm$1.13	&5.35$\pm$1.76	&1.95$\pm$1.67	&3.93$\pm$0.81	&5.24$\pm$1.00	&{\bf 0.96$\pm$0.72}	&1.41$\pm$0.66\\
\hline
\multicolumn{3}{c|}{Digits average} 
    & 14.62$\pm$0.80	&9.63$\pm$1.44	&9.50$\pm$1.51	&11.33$\pm$1.36	&27.80$\pm$0.75	&{\bf \textit{2.18$\pm$0.56}}	&{\bf 2.15$\pm$0.58} \\
\hline
\multirow{6}*{Office-31} & Amazon & DSLR
    &9.50$\pm$0.96	&1.99$\pm$1.60	&2.69$\pm$2.63	&1.49$\pm$1.38	&8.71$\pm$1.40	&1.85$\pm$1.22	&{\bf 0.96$\pm$0.87}\\
 & Amazon & Webcam
    &8.81$\pm$0.86	&{\bf 1.42$\pm$0.64}	&7.82$\pm$4.60	&2.19$\pm$1.35	&9.70$\pm$1.69	&4.43$\pm$2.10	&1.62$\pm$0.98\\
 & DSLR & Amazon
    &5.24$\pm$3.20	&10.47$\pm$1.81	&11.57$\pm$1.93	&4.21$\pm$2.76	&18.80$\pm$1.91	&9.05$\pm$1.78	&{\bf 3.28$\pm$1.97}\\
 & DSLR & Webcam
    &1.82$\pm$1.24	&0.74$\pm$0.88	&7.38$\pm$1.33	&{\bf 0.62$\pm$0.38}	&1.62$\pm$0.36	&2.97$\pm$0.52	&3.08$\pm$0.89\\
 & Webcam & Amazon
    &{\bf 3.07$\pm$2.26}	&11.98$\pm$3.05	&13.29$\pm$3.19	&7.39$\pm$3.63	&19.71$\pm$1.71	&9.56$\pm$1.77	&5.23$\pm$2.03\\
 & Webcam & DSLR
    &1.58$\pm$1.00	&1.22$\pm$1.09	&9.40$\pm$2.03	&0.50$\pm$0.26	&0.22$\pm$0.15	&{\bf 0.46$\pm$0.24}	&0.88$\pm$0.49\\
\hline
\multicolumn{3}{c|}{Office-31 average} 
    & 5.00$\pm$1.26	    &4.64$\pm$1.23	&8.69$\pm$1.62	&{\bf \textit{2.73$\pm$1.28}}	&9.79$\pm$1.10	&4.72$\pm$1.13	&{\bf {2.51$\pm$1.10}} \\
\hline
\multirow{12}*{Office-Home} & Art & Clipart
    &33.63$\pm$0.17	&10.03$\pm$2.21	&17.80$\pm$1.86	&3.88$\pm$2.07	&53.08$\pm$0.15	&8.79$\pm$0.24	&{\bf 3.41$\pm$0.43}\\
 & Art & Product
    &28.17$\pm$0.31	&6.91$\pm$1.31	&8.85$\pm$1.88	&2.80$\pm$1.37	&35.99$\pm$0.25	&3.83$\pm$0.38	&{\bf 0.86$\pm$0.24}\\
 & Art & Real-World
    &25.80$\pm$0.23	&6.77$\pm$2.01	&6.60$\pm$1.47	&1.91$\pm$1.13	&26.54$\pm$0.23	&5.75$\pm$0.63	&{\bf 1.26$\pm$0.57}\\
 & Clipart & Art
    &31.84$\pm$0.37	&11.38$\pm$2.49	&10.06$\pm$6.08	&1.85$\pm$0.99	&41.33$\pm$0.45	&9.32$\pm$0.63	&{\bf 6.02$\pm$0.74}\\
 & Clipart & Product
    &30.50$\pm$0.31	&9.75$\pm$2.91	&11.54$\pm$2.17	&{\bf 2.93$\pm$0.94}	&35.88$\pm$0.32	&12.59$\pm$0.34	&7.59$\pm$0.38\\
 & Clipart & Real-World
    &31.02$\pm$0.30	&9.94$\pm$3.13	&12.50$\pm$1.15	&{\bf 2.30$\pm$1.19}	&34.22$\pm$0.26	&11.94$\pm$0.41	&7.20$\pm$0.58\\
 & Product & Art
    &18.79$\pm$0.28	&12.51$\pm$2.41	&{\bf 6.56$\pm$4.04}	&6.94$\pm$2.23	&42.09$\pm$0.33	&10.51$\pm$0.83	&8.98$\pm$0.65\\
 & Product & Clipart
    &25.27$\pm$0.35	&12.15$\pm$2.56	&17.79$\pm$3.83	&{\bf 7.00$\pm$3.17}	&52.68$\pm$0.25	&17.12$\pm$0.23	&11.98$\pm$0.43\\
 & Product & Real-World
    &14.27$\pm$0.19	&6.33$\pm$1.58	&10.80$\pm$1.10	&{\bf 2.67$\pm$1.06}	&25.62$\pm$0.15	&7.00$\pm$0.31	&3.85$\pm$0.63\\
 & Real-World & Art
    &19.91$\pm$0.07	&6.59$\pm$0.84	&13.04$\pm$2.91	&4.98$\pm$1.50	&29.61$\pm$0.14	&5.44$\pm$0.42	&{\bf 3.26$\pm$0.41}\\
 & Real-World & Clipart
    &28.97$\pm$0.23	&8.80$\pm$0.98	&20.50$\pm$2.07	&{\bf 4.04$\pm$2.30}	&50.49$\pm$0.33	&8.96$\pm$0.38	&4.11$\pm$0.32\\
 & Real-World & Product
    &19.23$\pm$0.23	&3.79$\pm$1.28	&9.57$\pm$1.51	&2.61$\pm$0.73	&22.69$\pm$0.25	&3.54$\pm$0.32	&{\bf 1.28$\pm$0.15}\\
\hline
\multicolumn{3}{c|}{Office-Home average} 
    & 25.62$\pm$0.50	&8.75$\pm$1.41	&12.13$\pm$1.58	&{\bf 3.66$\pm$1.25}	&37.52$\pm$0.51	&8.73$\pm$0.65	&{\bf \textit{4.98$\pm$0.68}} \\
\hline
\multicolumn{3}{c|}{VisDA (Syn-to-Real)} 
    & 15.72$\pm$4.35	&8.90$\pm$1.71	&7.50$\pm$4.70	&{\bf \textit{4.41$\pm$2.52}}	&29.31$\pm$3.73	&{\bf \textit{4.41$\pm$1.10}}	&{\bf 1.73$\pm$0.93} \\
\Xhline{2\arrayrulewidth}
\end{tabular}
}

\vspace{-1ex}

\label{tab:full_result_natural} 
\end{table*}
%------------------------------------------------------------------------

%------------------------------------------------------------------------
\begin{table*}[t!]
\centering
\def\arraystretch{1.3}
\caption{ 
Full UAE benchmark results on synthetic distribution shift scenarios (MAE, \%). 
}
%\vspace{-1.4ex}
\resizebox{0.96\textwidth}{!}
{%
\begin{tabular}{c| c|c c c c c|c c}
\Xhline{2\arrayrulewidth}
\multicolumn{2}{c|}{}  & \multicolumn{5}{c|}{source access approach} & \multicolumn{2}{c}{source-free approach (ours)}  \\
\hline
datasets & corruptions
    & DoC \cite{ICCV2021_DoC}    & Proxy \cite{ICML2020_DIR}     & RI \cite{NEURIPS2021_RM} &RM \cite{NEURIPS2021_RM} & GDE \cite{ICLR2022_GDE}   
    & {SF-DAP (ADV)} & {SF-DAP (AAP)}\\

\hline
CIFAR-10 & Gaussian noise    
    &	22.33$\pm$3.33	&	11.20$\pm$0.31	&	3.79$\pm$1.95	&	{\bf 0.78$\pm$0.50}	&	4.86$\pm$1.48	&	1.87$\pm$0.43	&	3.08$\pm$2.33	\\
to & Shot noise        
    &	20.42$\pm$2.81	&	10.95$\pm$0.31	&	3.08$\pm$1.19	&	{\bf 0.52$\pm$0.30}	&	3.77$\pm$1.44	&	1.95$\pm$0.27	&	3.31$\pm$2.48	\\
CIFAR-10-C & Impulse noise
    &	27.94$\pm$5.23	&	12.25$\pm$1.04	&	4.97$\pm$2.35	&	2.81$\pm$1.00	&	8.01$\pm$2.03	&	{\bf 1.14$\pm$0.57}	&	3.37$\pm$2.05	\\
\multirow{16}*{} & Speckle noise
    &	20.49$\pm$2.81	&	10.93$\pm$0.23	&	3.12$\pm$1.14	&	{\bf 0.68$\pm$0.39}	&	3.68$\pm$1.41	&	1.79$\pm$0.31	&	3.04$\pm$2.21	\\
 & Defocus blur
    &	15.79$\pm$1.42	&	9.22$\pm$0.82	&	2.35$\pm$0.49	&	2.48$\pm$1.95	&	{\bf 1.64$\pm$1.20}	&	1.65$\pm$0.52	&	2.86$\pm$2.06	\\
 & Glass blur 
    &	19.26$\pm$0.51	&	11.96$\pm$1.19	&	{\bf 1.72$\pm$1.05}	&	2.19$\pm$1.69	&	2.62$\pm$1.97	&	2.20$\pm$0.43	&	3.18$\pm$2.66	\\
 & Motion blur
    &	18.96$\pm$1.62	&	10.93$\pm$0.83	&	2.70$\pm$0.29	&	2.58$\pm$1.88	&	2.16$\pm$1.36	&	{\bf 1.27$\pm$0.32}	&	3.38$\pm$1.68	\\
 & Zoom blur
    &	15.73$\pm$1.24	&	8.88$\pm$0.89	&	2.31$\pm$0.69	&	2.79$\pm$1.63	&	1.45$\pm$0.92	&	{\bf 1.43$\pm$0.81}	&	2.60$\pm$1.89	\\
 & Gaussian blur
    &	16.66$\pm$1.57	&	9.42$\pm$0.82	&	2.64$\pm$0.45	&	2.74$\pm$2.24	&	1.83$\pm$1.40	&	{\bf 1.55$\pm$0.45}	&	2.94$\pm$2.08	\\
 & Snow
    &	19.56$\pm$1.71	&	11.37$\pm$0.48	&	2.31$\pm$0.74	&	{\bf 1.39$\pm$0.55}	&	2.54$\pm$0.40	&	1.71$\pm$0.99	&	3.71$\pm$2.19	\\
 & Fog
    &	19.73$\pm$2.88	&	11.80$\pm$1.00	&	3.92$\pm$0.67	&	2.70$\pm$1.65	&	3.27$\pm$2.28	&	3.14$\pm$2.68	&	{\bf 2.56$\pm$2.35}	\\
 & Frost
    &	18.09$\pm$2.01	&	9.80$\pm$0.56	&	2.81$\pm$0.43	&	2.02$\pm$0.76	&	2.35$\pm$0.71	&	{\bf 1.40$\pm$0.83}	&	2.34$\pm$1.85	\\
 & Brightness
    &	15.17$\pm$2.08	&	8.69$\pm$0.49	&	2.53$\pm$0.95	&	2.26$\pm$1.88	&	1.75$\pm$1.17	&	{\bf 1.29$\pm$0.40}	&	2.79$\pm$1.93	\\
 & Spatter
    &	19.78$\pm$3.08	&	11.48$\pm$0.31	&	3.16$\pm$0.61	&	{\bf 1.35$\pm$1.05}	&	2.66$\pm$1.62	&	2.21$\pm$0.85	&	3.44$\pm$2.76	\\
 & Contrast
    &	23.34$\pm$6.17	&	9.70$\pm$1.09	&	8.86$\pm$1.72	&	4.30$\pm$2.72	&	6.65$\pm$4.98	&	11.66$\pm$5.42	&	{\bf 2.90$\pm$0.59}	\\
 & Elastic transform 
    &	16.60$\pm$1.30	&	10.63$\pm$1.28	&	2.11$\pm$0.26	&	2.78$\pm$1.33	&	{\bf 1.33$\pm$0.91}	&	1.87$\pm$0.53	&	2.96$\pm$2.44	\\
 & Pixelate
    &	15.34$\pm$1.25	&	9.16$\pm$0.73	&	1.92$\pm$0.34	&	2.34$\pm$1.04	&	{\bf 1.19$\pm$0.60}	&	1.77$\pm$0.36	&	2.82$\pm$2.10	\\
 & JPEG compression 
    &	16.82$\pm$1.77	&	10.29$\pm$0.37	&	2.60$\pm$0.25	&	{\bf 1.78$\pm$1.52}	&	2.02$\pm$1.48	&	2.09$\pm$0.96	&	3.34$\pm$2.32	\\
 & Saturate
    &	17.30$\pm$3.34	&	10.34$\pm$0.33	&	3.83$\pm$1.25	&	2.70$\pm$2.39	&	3.06$\pm$1.90	&	{\bf 2.39$\pm$2.19}	&	3.56$\pm$2.55	\\
\hline
\multicolumn{2}{c|}{CIFAR-10 to CIFAR-10-C average} 
    & 18.91$\pm$1.56	&10.47$\pm$0.83	&3.20$\pm$0.94	&{\bf 2.17$\pm$1.18}	&2.99$\pm$1.24	&{\bf \textit{2.34$\pm$1.47}}	&{ 3.06$\pm$1.46} \\
\hline
CIFAR-100 & Gaussian noise    
    &	52.05$\pm$6.07	&	22.92$\pm$13.56	&	7.38$\pm$4.02	&	{\bf 0.70$\pm$0.50}	&	7.97$\pm$3.11	&	2.98$\pm$1.69	&	2.48$\pm$1.92	\\
to & Shot noise        
    &	50.05$\pm$5.24	&	19.20$\pm$2.71	&	6.52$\pm$2.89	&	{\bf 1.17$\pm$0.24}	&	6.74$\pm$3.08	&	4.30$\pm$2.17	&	2.88$\pm$2.51	\\
CIFAR-100-C & Impulse noise
    &	55.92$\pm$9.18	&	39.79$\pm$27.47	&	9.39$\pm$5.13	&	{\bf 0.55$\pm$0.39}	&	10.17$\pm$4.00	&	0.70$\pm$0.38	&	1.52$\pm$0.31	\\
\multirow{16}*{} & Speckle noise
    &	50.30$\pm$5.30	&	19.49$\pm$2.58	&	6.54$\pm$2.80	&	{\bf 1.16$\pm$0.32}	&	6.73$\pm$3.11	&	4.32$\pm$2.31	&	2.70$\pm$2.20	\\
 & Defocus blur
    &	45.08$\pm$2.52	&	30.47$\pm$6.06	&	4.93$\pm$1.40	&	{\bf 2.61$\pm$1.25}	&	4.87$\pm$3.13	&	9.95$\pm$3.01	&	3.06$\pm$1.71	\\
 & Glass blur 
    &	48.00$\pm$1.40	&	24.54$\pm$0.79	&	3.50$\pm$3.10	&	{\bf 0.91$\pm$0.40}	&	5.61$\pm$1.47	&	7.07$\pm$4.21	&	2.86$\pm$0.88	\\
 & Motion blur
    &	47.36$\pm$2.35	&	30.12$\pm$7.08	&	4.73$\pm$2.31	&	{\bf 2.17$\pm$0.51}	&	5.56$\pm$2.54	&	9.27$\pm$2.32	&	3.83$\pm$2.58	\\
 & Zoom blur
    &	44.92$\pm$2.34	&	29.26$\pm$6.25	&	4.80$\pm$1.23	&	{\bf 2.50$\pm$1.55}	&	4.45$\pm$2.89	&	8.80$\pm$2.46	&	2.89$\pm$2.54	\\
 & Gaussian blur
    &	46.09$\pm$2.53	&	30.44$\pm$6.48	&	5.04$\pm$1.76	&	{\bf 2.60$\pm$0.88}	&	5.22$\pm$3.25	&	10.12$\pm$2.56	&	3.17$\pm$2.93	\\
 & Snow
    &	49.93$\pm$3.06	&	22.15$\pm$0.48	&	4.60$\pm$3.65	&	{\bf 1.02$\pm$0.33}	&	6.39$\pm$1.16	&	6.32$\pm$2.22	&	1.76$\pm$1.06	\\
 & Fog
    &	51.15$\pm$3.62	&	31.95$\pm$9.00	&	5.58$\pm$3.96	&	2.49$\pm$0.73	&	7.48$\pm$3.29	&	6.51$\pm$2.77	&	{\bf 1.81$\pm$1.03}	\\
 & Frost
    &	48.58$\pm$3.36	&	24.23$\pm$10.16	&	4.77$\pm$3.19	&	{\bf 1.70$\pm$0.28}	&	6.26$\pm$1.54	&	4.03$\pm$1.92	&	{\bf 1.70$\pm$1.16}	\\
 & Brightness
    &	45.06$\pm$3.61	&	23.15$\pm$1.78	&	5.27$\pm$1.52	&	{\bf 2.47$\pm$1.12}	&	4.78$\pm$3.13	&	7.86$\pm$2.65	&	3.71$\pm$2.92	\\
 & Spatter
    &	50.44$\pm$4.65	&	20.59$\pm$1.32	&	5.98$\pm$3.33	&	{\bf 1.43$\pm$0.22}	&	6.65$\pm$3.45	&	5.53$\pm$2.33	&	2.89$\pm$2.05	\\
 & Contrast
    &	52.17$\pm$5.73	&	24.20$\pm$13.17	&	10.12$\pm$4.47	&	3.03$\pm$0.52	&	9.51$\pm$6.13	&	{\bf 2.18$\pm$0.92}	&	11.03$\pm$2.70	\\
 & Elastic transform 
    &	45.63$\pm$2.60	&	29.04$\pm$1.92	&	4.57$\pm$1.99	&	{\bf 2.10$\pm$1.51}	&	4.74$\pm$2.34	&	9.61$\pm$3.82	&	2.97$\pm$0.49	\\
 & Pixelate
    &	44.42$\pm$2.66	&	25.03$\pm$1.54	&	4.42$\pm$1.67	&	{\bf 1.99$\pm$1.39}	&	4.58$\pm$2.29	&	8.43$\pm$3.69	&	2.87$\pm$1.10	\\
 & JPEG compression 
    &	46.85$\pm$3.15	&	26.38$\pm$0.75	&	5.05$\pm$1.87	&	2.41$\pm$1.03	&	5.19$\pm$3.15	&	8.62$\pm$3.34	&	{\bf 1.93$\pm$1.22}	\\
 & Saturate
    &	50.20$\pm$5.78	&	27.58$\pm$1.91	&	7.68$\pm$2.63	&	{\bf 3.17$\pm$0.22}	&	7.29$\pm$4.97	&	12.80$\pm$3.36	&	6.31$\pm$3.47	\\
\hline
\multicolumn{2}{c|}{CIFAR-100 to CIFAR-100-C average} 
    & 48.64$\pm$1.99	&26.34$\pm$2.46	&5.84$\pm$1.67	&{\bf 1.90$\pm$0.84}	&6.33$\pm$1.75	&6.81$\pm$1.59	&{\bf \textit{3.28$\pm$1.35}} \\
\Xhline{2\arrayrulewidth}
\end{tabular}
}

\vspace{-1ex}

\label{tab:full_result_synthetic} 
\end{table*}
%------------------------------------------------------------------------

\subsection{Training Configurations}
\paragraph{Source Model Training.}
During the source model training, we train the models from the scratch for the smaller networks such as the LeNet variant (Digits) and ResNet18 (CIFAR-10 and CIFAR-100). We leverage ImageNet pre-trained initialization for ResNet50 (Office-31 and Office-Home) and ResNet101 (VisDA) to compensate for the small dataset size and to reduce time to converge, respectively.

Referring to the {\it random initialization} strategy of \cite{ICLR2022_GDE} combined by additional hyperparameter selection,
we train ten source models for each source dataset independently, 
by applying different augmentations out of \{weak augmentation, strong augmentation\} to input, different initial seeds for random number generation selected from \{2021, 2022, 2023, 2024, 2025\}, and different learning rates if necessary.
Note that MNIST and USPS are trained only with strong augmentation as the challenging M$\rightarrow$S and U$\rightarrow$S UDA on source models trained with weak augmentation often lead to collapsed results.
Specifically, standard random cropping, rotating, flipping, and color jittering are applied for weak augmentation (on Digits, flipping is not used) whereas we employ RandAugment \cite{RandAugment} with Cutout \cite{Cutout} for strong augmentation. 

Each source model of Digits is trained for 50 epochs with a mini-batch size of 500 while the learning rate is initially set to one of \{0.1, 0.05\} and steps down by $\times$0.1 after 40-th epoch. 
For CIFAR-10 and CIFAR-100, we train from scratch for 300 epochs with a mini-batch size of 200 while the learning rate is initially set to 0.1 and steps down by $\times$0.1 every 80-th epoch. 
We fine-tune the source models of Office-31, Office-Home, and VisDA using ImageNet pre-trained backbone with learning rate schedules ranging from 0.001 to 0.00001 on cosine annealing for 1200, 1200, and 400 iterations, respectively, with reference to a recent model training tricks \cite{Tricks}.
We set the mini-batch size to 192, 192, and 132 for Office-31, Office-Home, and VisDA, respectively.
For CIFAR-10 and CIFAR-100, we apply Adam \cite{Adam} optimizer whereas SGD with momentum 0.9 and weight decay 0.0001 is applied for the other models.

During investigation on straightforward pseudo-labeling (Sec 3 in the main manuscript), we trained 192 ResNet50 source models by combining six different learning rates, two input augmentation options, two optimizers among \{Adam, SGD with momentum\}, two label smoothing options and four different training epochs.

\paragraph{Source-Free UDA.}
In the proposed SF-DAP framework, we first adapt the given source model to the target distribution by a source-free UDA method such as SHOT \cite{SHOT}, FAUST \cite{FAUST}, or the proposed PAFA.
We apply the default hyperparameters that are used by each method unless otherwise stated.
The same random seeds are applied that are used for the source model training.
Two types of augmentations are employed to enable perturbation in PAFA: standard random crop-rotate-flip and color jittering for weak augmentation (no flipping is used on Digits), and RandAugment \cite{RandAugment} with Cutout \cite{Cutout} for strong augmentation. 
In PAFA, we simply set $\alpha$ to 0.5 without any intensive tuning effort since this value leads to a viable performance.

In all source-free UDA methods, the common training configurations are as follows:
For Digits and Office-31 settings, we apply a fixed learning rate of 0.0002. 
VisDA is trained by a learning rate from 0.0005 to 0.00005 scheduled with cosine annealing, whereas a learning rate from 0.001 to 0.0001 scheduled with cosine annealing is applied to CIFAR-10$\rightarrow$CIFAR-10-C and CIFAR-100$\rightarrow$CIFAR-100-C scenarios.
Mini-batch size is set to 500 in Digits, CIFAR-10-C, and CIFAR-100-C target domains while 92 in Office-31 and 64 in VisDA are applied.
The determination of the number of UDA epochs is based on the point of loss saturation within 60 epochs for Digits, Office-31, and Office-Home datasets, and 4, 20, and 20 epochs for VisDA, CIFAR-10-C, and CIFAR-100-C datasets, respectively.

\subsection{Evaluation Protocol}
We repreat experiments on ten source models trained independently and present the mean and standard deviation of the mean absolute errors (MAE) for 63 scenarios distributed across six groups. 
Given the variability in the number of tasks within each group, we computed two types of averages to provide a comprehensive evaluation of the estimation performance across benchmark groups. 
The first average, referred to as the {\it micro average}, represents the overall average. 
The second average, presented as the {\it macro average}, considers the mean of the average MAE values within each group, allowing for a more balanced assessment of the estimation performance.

We evaluate each baseline according to the definition of the risk estimator as reported in each paper.
We utilize $\E_{x\sim\D_S}[\max_{k\in\Y}h_S^{k}(x)] - \E_{x\sim\D_T}[\max_{k\in\Y}h_S^{k}(x)]$ for DoC \cite{ICCV2021_DoC}, whereas $\max_{h'\in\P}~\varepsilon_T(h_S, h')$ is used for Proxy \cite{ICML2020_DIR} where $\P$ is a set of check models. 
For GDE \cite{ICLR2022_GDE}, we apply $\E_{h_S,h'_S\in\S}[\varepsilon_T(h_S(x), h'_S(x))]$ where $\S$ is a set of sibling source models that includes the source model of interest for evaluation.
For RI and RM \cite{NEURIPS2021_RM}, iterative self-trained ensemble of models $\{h_i\}^N_{i=1}$ as a pseudo-label is used to estimate $\varepsilon_T(h_S)$.
For a fair comparison, five pairs of sibling source models are used for GDE experiments and five independently-adapted check models are used for Proxy Risk experiments.
We did our best to achieve the reported results, but the higher MAE numbers, if any, may be partially due to our lack of a proper recipe for adjusting hyperparameters when running baseline approaches.

\subsection{Detailed Results}
All results of 63 scenarios are presented in Table \ref{tab:full_result_natural} for the natural distribution shift scenarios and in Table \ref{tab:full_result_synthetic} for the synthetic distribution shift scenarios.

\subsection{Computing Infrastructure}
We conduct all experiments using PyTorch and NVIDIA A100 GPUs.

%------------------------------------------------------------------------
\section{Review on Existing UAE Methods}
\label{app:methods}

\subsection{Regression-based Approaches}
To estimate the accuracy of an already trained model to unknown unlabeled data, one of the first approaches \cite{CVPR2021_Deng} draw on the (negative) correlation between the differences in data distribution and model accuracy, and builds a regression model from the Frechet distance between a collection of artificially augmented from the source data and the model accuracy on these augmented datasets. \cite{ICML2021_Deng} established the relation between the rotation estimation task and the classification task. When the network is trained to optimize both criteria, they observed a strong correlation between rotation estimation accuracy and classification accuracy. This observation led to a simple regression-based classification accuracy estimation approach by fitting a linear mapping between both accuracies from a collection of datasets synthetically generated from standard datasets
\cite{CVPR2021_Deng}.

\subsection{Confidence-based Approaches}
Guillory \etal \cite{ICCV2021_DoC} approaches the problem of accuracy estimation from the per-sample prediction confidence that is robust to distributional shifts. They came up with a simple prediction confidence-based measure called AC (Average Confidence) and extended it to DoC (Difference of Confidence) measure, which shows improved estimation accuracy. In a similar manner, \cite{ICLR2022_ATC} developed a ‘score function’ based accuracy estimation that is essentially per-sample confidence measures from either maximum class probability or negative entropy of the class probability. They first calibrate the probability outputs so that they match the accuracy using temperature scaling \cite{Calibration_ICML17}. Then the key to their approach is to identify the score function threshold from the source data that match the source accuracy and apply the same threshold to the target score function leading to the target accuracy estimation.

\subsection{Disagreement-based Approaches}
Nakkiran and Bansal \cite{Nakkiran_Bansal} first found that if one train two networks of identical architecture on two independently sampled subsets of a dataset, the disagreement rate on test data linearly correlates with the network’s test accuracy. 
Jiang \etal \cite{ICLR2022_GDE} further extends the behavior to two identical networks trained on the same dataset but with different random initialization. They verified that this observed correlation leads to a {\it disagreement} based accuracy estimation for unlabeled datasets, called GDE. They further established the necessary conditions for the calibration of the prediction outputs for the method to work.

%------------------------------------------------------------------------
\begin{table*}[!t]
\centering
\def\arraystretch{1.1}
\caption{ 
Ablation study on various UAE tasks by comparing SF-DAP (ADV), SF-DAP (AAP) and their intermediate configuration.
We report average MAEs(\%) in each benchmark group for simplicity.
The notation $C_{adj\_unc}C_{cls}$ refers to an intermediate configuration that is identical to SF-DAP (AAP) except that the data volume density factor $C_{den}$ is not used.
Micro average computes the mean of all 63 cross-domain scenarios, whereas macro average represents the mean of average MAE values within the six benchmark groups.
}
%\vspace{-1.4ex}
\resizebox{0.86\textwidth}{!}
{%
\begin{tabular}{c | c c c c c c | c| c }
\Xhline{2\arrayrulewidth}
Method 
    & Digits & Office-31 & Office-Home
    & VisDA & CIFAR-10-C & CIFAR-100-C & micro avg. & macro avg.\\
% \midrule
\hline
SF-DAP (ADV)
    &2.18$\pm$0.56	&4.72$\pm$1.13	&8.73$\pm$0.65	&4.41$\pm$1.10	&2.34$\pm$1.01	&6.81$\pm$1.59	&5.15$\pm$1.15	&4.86$\pm$1.00\\
$C_{adj\_unc}C_{cls}$
    &{\bf 2.05$\pm$0.61}	&3.05$\pm$1.07	&6.44$\pm$1.75	&3.05$\pm$0.86	&4.07$\pm$1.47	&3.83$\pm$1.33	&4.15$\pm$1.39	&3.75$\pm$1.09\\
SF-DAP (AAP)
    &2.15$\pm$0.58	&{\bf 2.51$\pm$1.10}	&{\bf 4.98$\pm$0.68}	&{\bf 1.73$\pm$0.93}	&{\bf 3.06$\pm$1.46}	&{\bf 3.28$\pm$1.35}	&{\bf 3.33$\pm$1.20}	&{\bf 2.95$\pm$1.01}\\

\Xhline{2\arrayrulewidth}
\end{tabular}
}%
\vspace{-1ex}
\label{tab:ablation_0}
\end{table*}
%------------------------------------------------------------------------

%------------------------------------------------------------------------
% \begin{table}
\begin{table*}[!t]
\centering
\def\arraystretch{1.1}
\caption{ 
Running time comparison of various methods that require additional training. 
For GDE and RI, the time for additional training of the source models is excluded.
}
%\vspace{-1.4ex}
\resizebox{0.70\textwidth}{!}
{%
\begin{tabular}{c | c c c c | c c}
\Xhline{2\arrayrulewidth}
Setting 
    & GDE \cite{ICLR2022_GDE} & RI \cite{NEURIPS2021_RM} & Proxy \cite{ICML2020_DIR} 
    & RM \cite{NEURIPS2021_RM} & SF-DAP (ADV) & SF-DAP (AAP) \\
\hline
Amazon$\rightarrow$DSLR     
    &27s   &1m 24s   &19m 3s   &18m 58s  &5m 54s   &6m 21s   \\
Amazon$\rightarrow$Webcam      
    &48s   &2m 35s   &30m 25s  &30m 50s  &9m 58s   &10m 49s   \\
CIFAR-10 (average)
    &13s   &1m 20s   &6m 43s   &5m 30s   &4m 40s   &4m 55s   \\
\Xhline{2\arrayrulewidth}
\end{tabular}
}%
\vspace{-1ex}
\label{tab:run_time}
\end{table*}
%------------------------------------------------------------------------

\subsection{Iterative Ensemble-based Approaches}
Chen \etal \cite{NEURIPS2021_RM} also extends the disagreement-based approach with the flavor of the UDA approach. It employs model ensembles of more than two, where the generated ensemble serves as the {\it check model} (by a majority vote) for the given source model to be evaluated. 
The ensembles are generated by either (1) random initialization (RI), or (2) random checkpoints when the models are trained to match the source and target features (RM), as in the well-known domain adaptation training \cite{DANN}. 
The approach also improves accuracy by using pseudo labels generated from the disagreement to be fed back to the training. 
A slightly earlier work \cite{ICML2020_DIR} also uses DIR training to find a check model that shows maximum disagreement while minimizing the DIR loss. 
Then the disagreement rate (called Proxy Risk) approximates the error of the given source model on the target data.

\section{Further Analysis}
\label{app:analysis}
\subsection{Ablation Study}
We conducted an ablation study on various datasets by comparing SF-DAP (AAP), SF-DAP (ADV), and their intermediate configuration as shown in Table \ref{tab:ablation_0}, where the magnitude of VAP ($\epsilon$) is computed by $\epsilon_0 C_{adj\_unc} C_{cls}$ for the intermediate configuration. 
Table \ref{tab:ablation_0} demonstrates the gradual improvement of estimation performance from SF-DAP (ADV) to SF-DAP (AAP).

%------------------------------------------------------------------------
\begin{table}[!t]
\centering
\def\arraystretch{1.3}
\caption{ 
UAE performance comparison when different UDA methods are employed. 
$^*$SHOT-IM is applied without the network augmentation.
$^{**}$Epistemic uncertainty loss is not applied.
}
%\vspace{-1.4ex}
\resizebox{0.92\columnwidth}{!}
{%
\begin{tabular}{c| c c c}
\Xhline{2\arrayrulewidth}
datasets 
    & PAFA   & SHOT$^*$ & FAUST$^{**}$\\
\hline
Digits 
    &{\bf{2.15$\pm$0.58}}	&3.33$\pm$1.63	&2.95$\pm$1.18\\
Office-31 
    &{\bf{2.51$\pm$1.10}}	&3.85$\pm$1.60	&3.79$\pm$1.54\\
 Office-Home 
    &{\bf{4.98$\pm$0.68}}	&5.67$\pm$1.72	&5.75$\pm$1.66\\
VisDA 
    &1.73$\pm$0.93	&{\bf{1.10$\pm$1.01}}	&1.16$\pm$0.95\\
\hline
CIFAR-10 
    &{\bf{3.06$\pm$1.46}}	&4.02$\pm$1.45	&3.95$\pm$1.46\\
CIFAR-100 
    &{\bf{3.28$\pm$1.35}}	&3.34$\pm$1.31	&3.39$\pm$1.33\\
\hline
micro average
    &{\bf{3.33$\pm$1.20}}	&4.00$\pm$1.49	&3.97$\pm$1.44\\
\hline
macro average 
    &{\bf{2.95$\pm$1.01}}	&3.55$\pm$1.21	&3.50$\pm$1.16\\
\Xhline{2\arrayrulewidth}
\end{tabular}
}

\vspace{-1ex}

\label{tab:different_uda} 
\end{table}
%------------------------------------------------------------------------

% ------------------------------------------------------------------------------
\subsection{Estimation Time}
\label{app:run_time}
We compare the runtime of various methods that require additional training as shown in Table \ref{tab:run_time}.
SF-DAP shows comparable or superior running time to other existing methods, particularly when compared with RM and Proxy Risk, which also perform UDA during estimation.
Once the target model adaptation is completed, the inference time of SF-DAP should be almost the same as that of RI.
Note that these results are measured with A100 GPUs.

\subsection{Performance with Different UDA Methods}
We evaluate some other source-free UDA methods such as SHOT-IM \cite{SHOT} and FAUST \cite{FAUST} within the proposed framework.
As shown in Table \ref{tab:different_uda}, there are no significant differences between their results, while PAFA shows the most preferable performance in our SF-DAP framework under both natural and synthetic distribution shift scenarios. 
Because of these superior results, we consider PAFA as our UDA recommendation for the proposed framework.
However, the potential benefits of more diverse source-free UDA methods are worth exploring in future research that jointly tune the feature generator with the head classifier.

% ------------------------------------------------------------------------------
\subsection{Uncertainty Measurement}
\label{app:uncertainty}
Recent studies have shown the dropout during inference, known as Monte Carlo (MC) dropout sampling, is equivalent to an approximation of a deep Gaussian process \cite{Gal16}. 
In this work, we estimate the predictive uncertainty by using the standard deviation of multiple stochastic forward passes that leverage MC dropout.
To enhance the accuracy estimation further, future research may explore the utilization of newly developed uncertainty measures, such as the balanced entropy, which captures the information balance between the model and the class label suggested by \cite{jaeoh_iclr}.

%------------------------------------------------------------------------
\begin{table}[!t]
\centering
\def\arraystretch{1.3}
\caption{ 
UAE performance comparison. 
RND, ADV, and AAP denote the perturbation method of the SF-DAP framework.
RND (ens.) uses the ensemble of the five independent RND estimates.
}
%\vspace{-1.4ex}
\resizebox{1.0\columnwidth}{!}
{%
\begin{tabular}{c| c c c c}
\Xhline{2\arrayrulewidth}
datasets 
    & RND   & RND (ens.) & ADV & AAP\\
\hline
Digits 
    &4.00$\pm$0.51	&{\bf{0.94$\pm$0.50}}	&2.18$\pm$0.56	&2.15$\pm$0.58\\
Office-31 
    &5.93$\pm$1.64	&3.58$\pm$1.14	&4.72$\pm$1.13	&{\bf{2.51$\pm$1.10}}\\
Office-Home 
    &{\bf{2.19$\pm$0.70}}	&9.03$\pm$0.66	&8.73$\pm$0.65	&4.98$\pm$0.68\\
VisDA 
    &{\bf{0.85$\pm$0.64}}	&7.14$\pm$1.25	&4.41$\pm$1.10	&1.73$\pm$0.93\\
\hline
CIFAR-10 
    &20.83$\pm$2.58	&4.32$\pm$1.90	&{\bf{2.34$\pm$1.01}}	&3.06$\pm$1.46\\
CIFAR-100 
    &17.57$\pm$2.09	&3.97$\pm$1.39	&6.81$\pm$1.59	&{\bf{3.28$\pm$1.35}}\\
\hline
micro average
    &12.96$\pm$1.92	&4.77$\pm$1.38	&5.15$\pm$1.15	&{\bf{3.33$\pm$1.20}}\\
\hline
macro average 
    &8.56$\pm$1.17	&4.83$\pm$1.07	&4.86$\pm$1.00	&{\bf{2.95$\pm$1.01}}\\
\Xhline{2\arrayrulewidth}
\end{tabular}
}

\vspace{-1ex}

\label{tab:rnd_ensemble} 
\end{table}
%------------------------------------------------------------------------

\subsection{Ensemble Effect on SF-DAP (RND)}
Applying an ensemble to SF-DAP (RND) yields estimation performance similar to or better than that of SF-DAP (ADV) as presented in Table \ref{tab:rnd_ensemble}.

As we have shown from RI and RM methods in Table \ref{tab:full_result_natural} and Table \ref{tab:full_result_synthetic}, the ensemble approach notably improves the accuracy estimation performance in general which is comparable with the intermediate configuration $C_{adj\_unc}C_{cls}$ shown in Table \ref{tab:ablation_0}. We interpret that the ensemble approach can equate to the role of capturing model uncertainty. However, our consolidated SF-DAP (AAP) shows superior performance and requires fewer computational resources compared to the ensemble method.

% -----------------------------------------------------------------------

\subsection{UAE Performance during UDA}
\begin{figure*}[!t]
  \centering
  \begin{minipage}[t]{0.28\textwidth} \centering \footnotesize
  \includegraphics[width=0.95\linewidth]{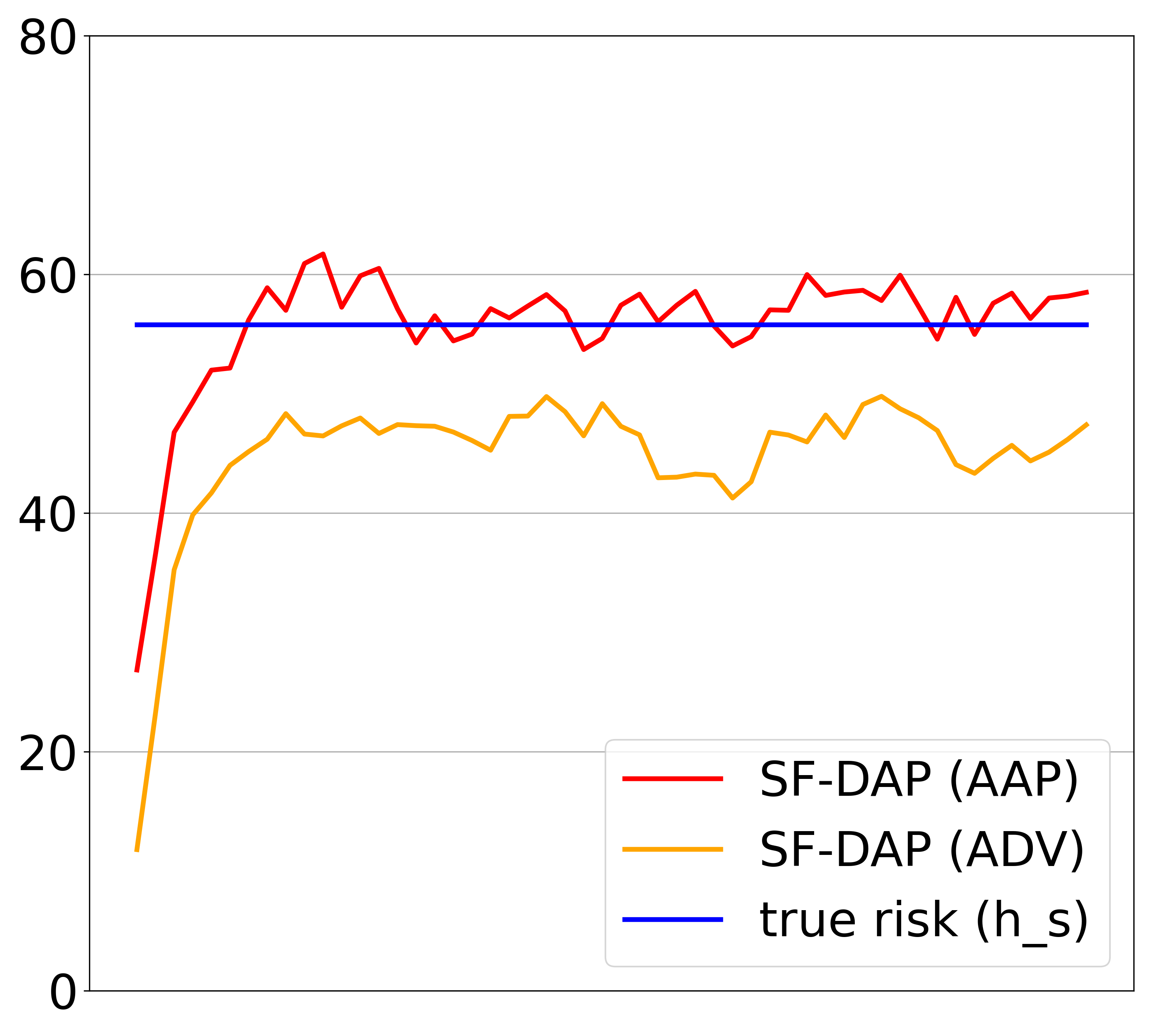}
  (a) Ar$\rightarrow$Cl
  \end{minipage}~
  \begin{minipage}[t]{0.28\textwidth} \centering \footnotesize
  \includegraphics[width=0.95\linewidth]{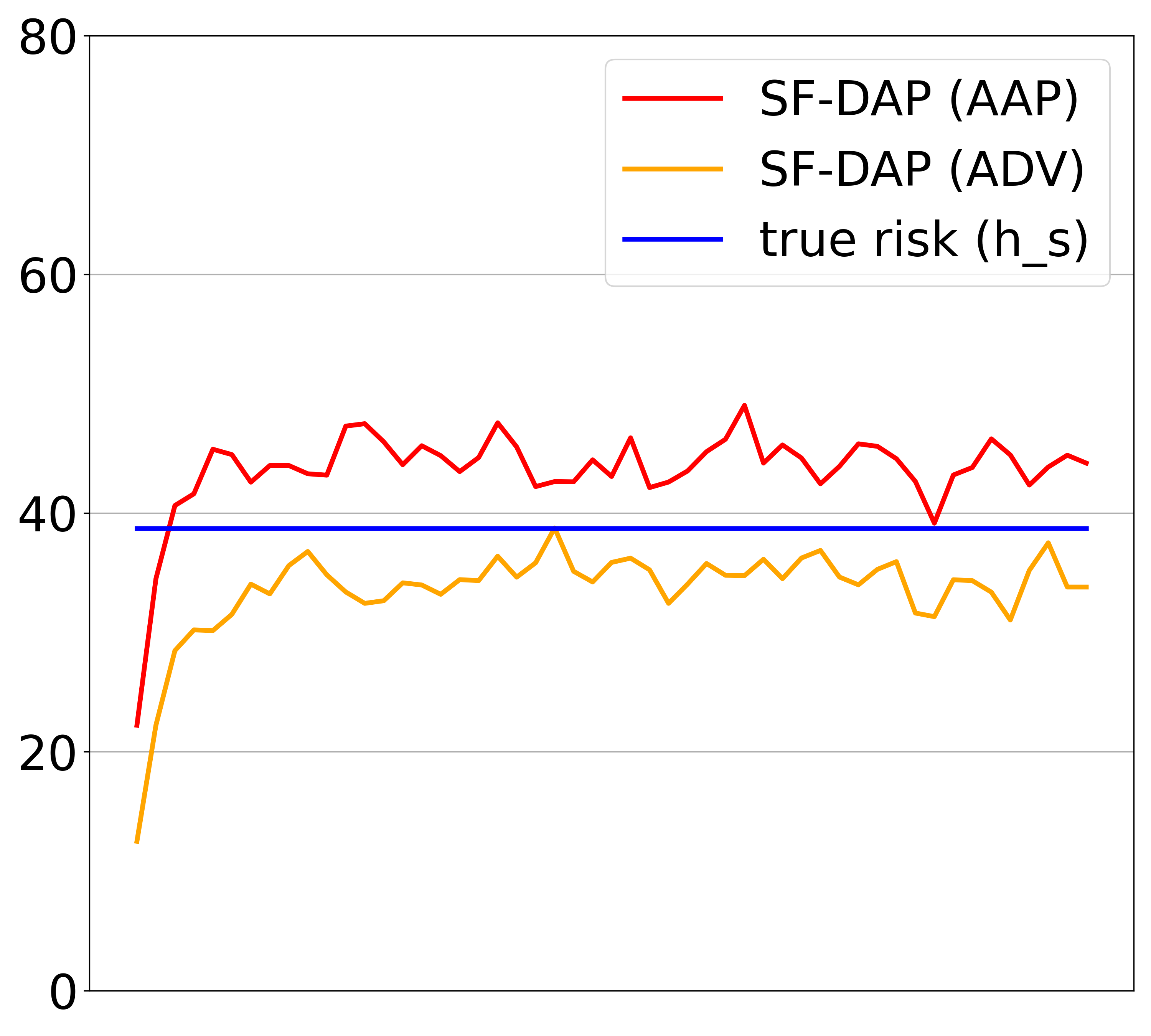}
  (b) Ar$\rightarrow$Pr
  \end{minipage}~
  \begin{minipage}[t]{0.28\textwidth} \centering \footnotesize
  \includegraphics[width=0.95\linewidth]{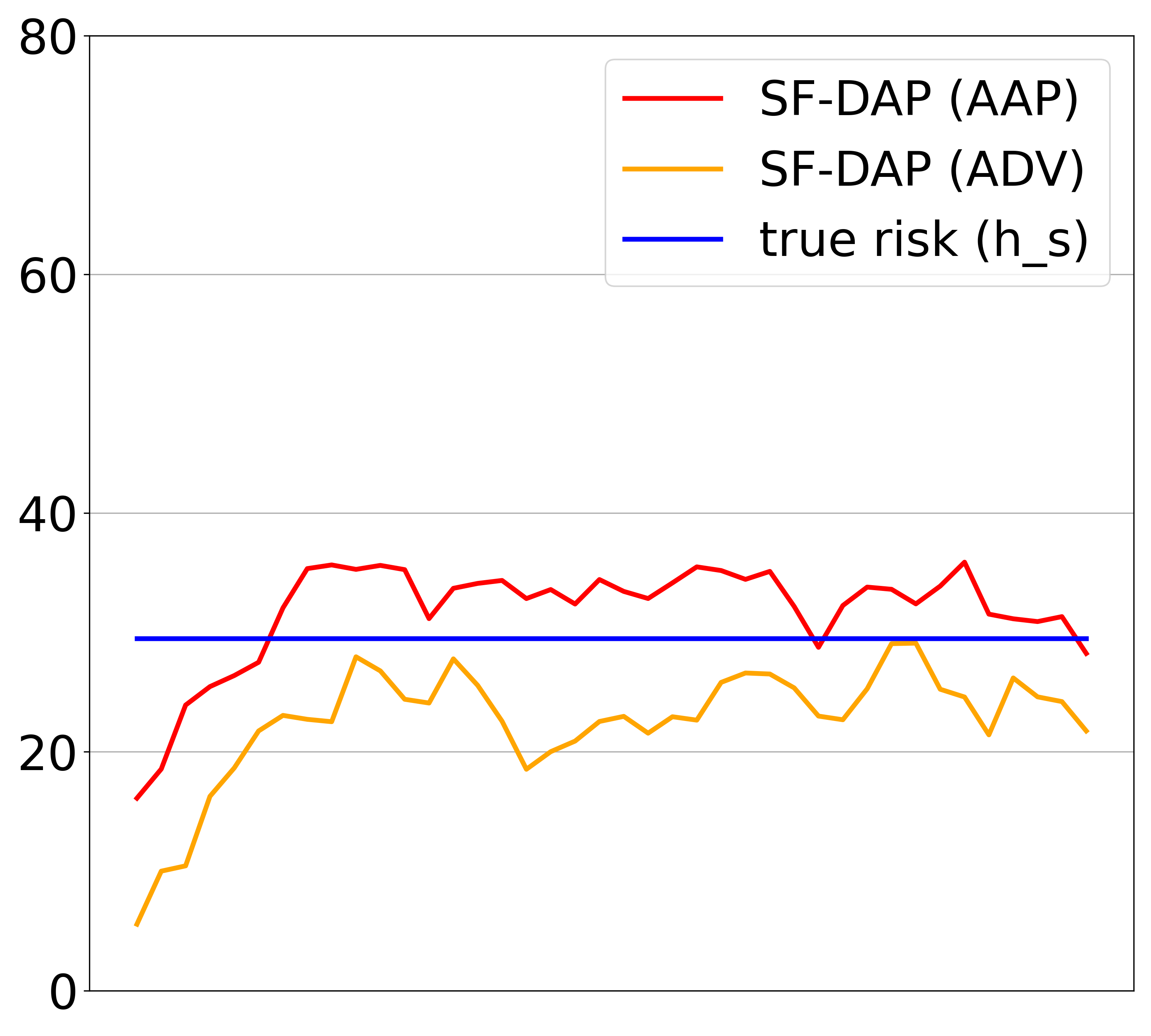}
  (c) Ar$\rightarrow$Re
  \end{minipage}\\
  \begin{minipage}[t]{0.28\textwidth} \centering \footnotesize
  \includegraphics[width=0.95\linewidth]{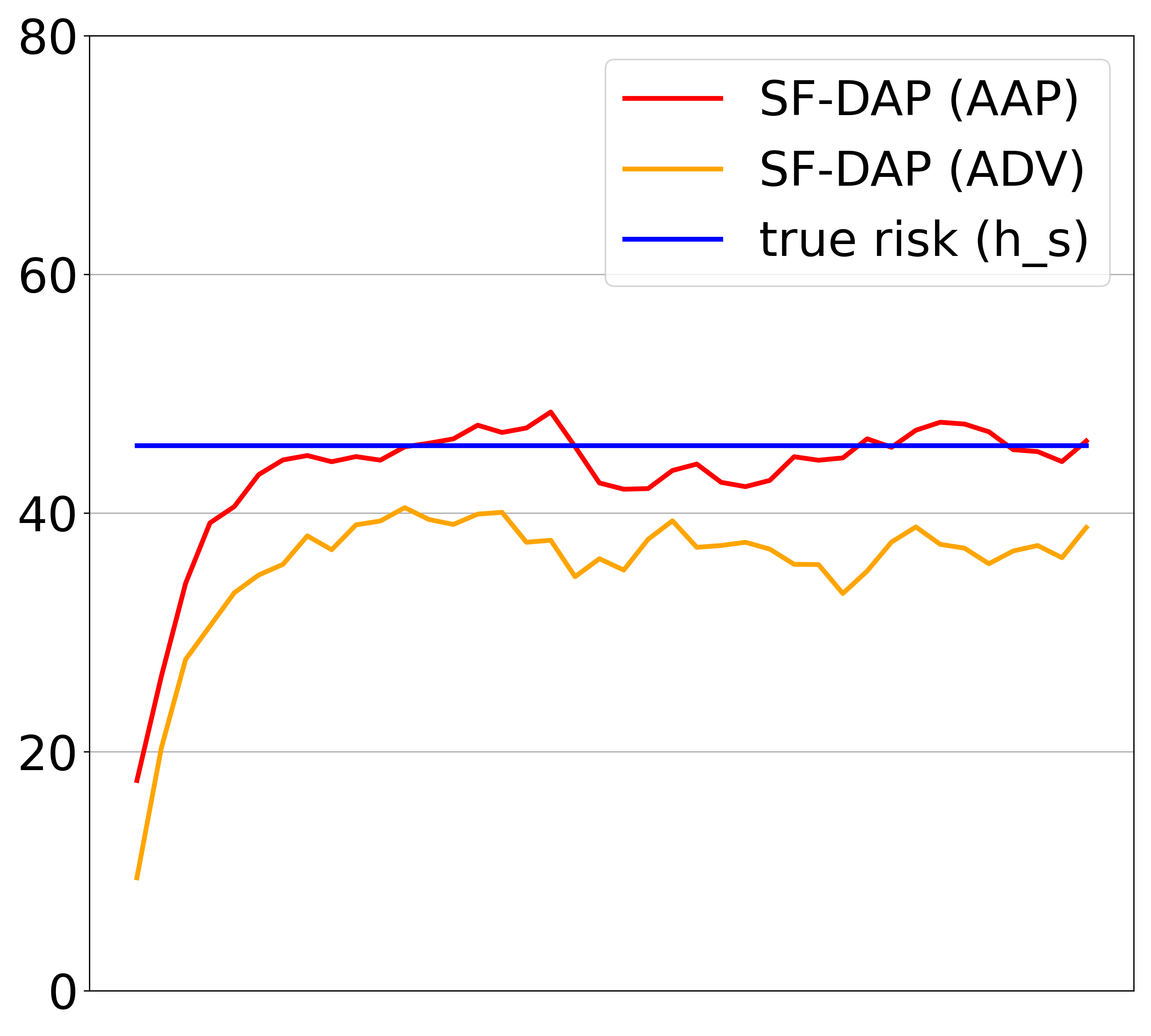}
  (d) Cl$\rightarrow$Ar
  \end{minipage}~
  \begin{minipage}[t]{0.28\textwidth} \centering \footnotesize
  \includegraphics[width=0.95\linewidth]{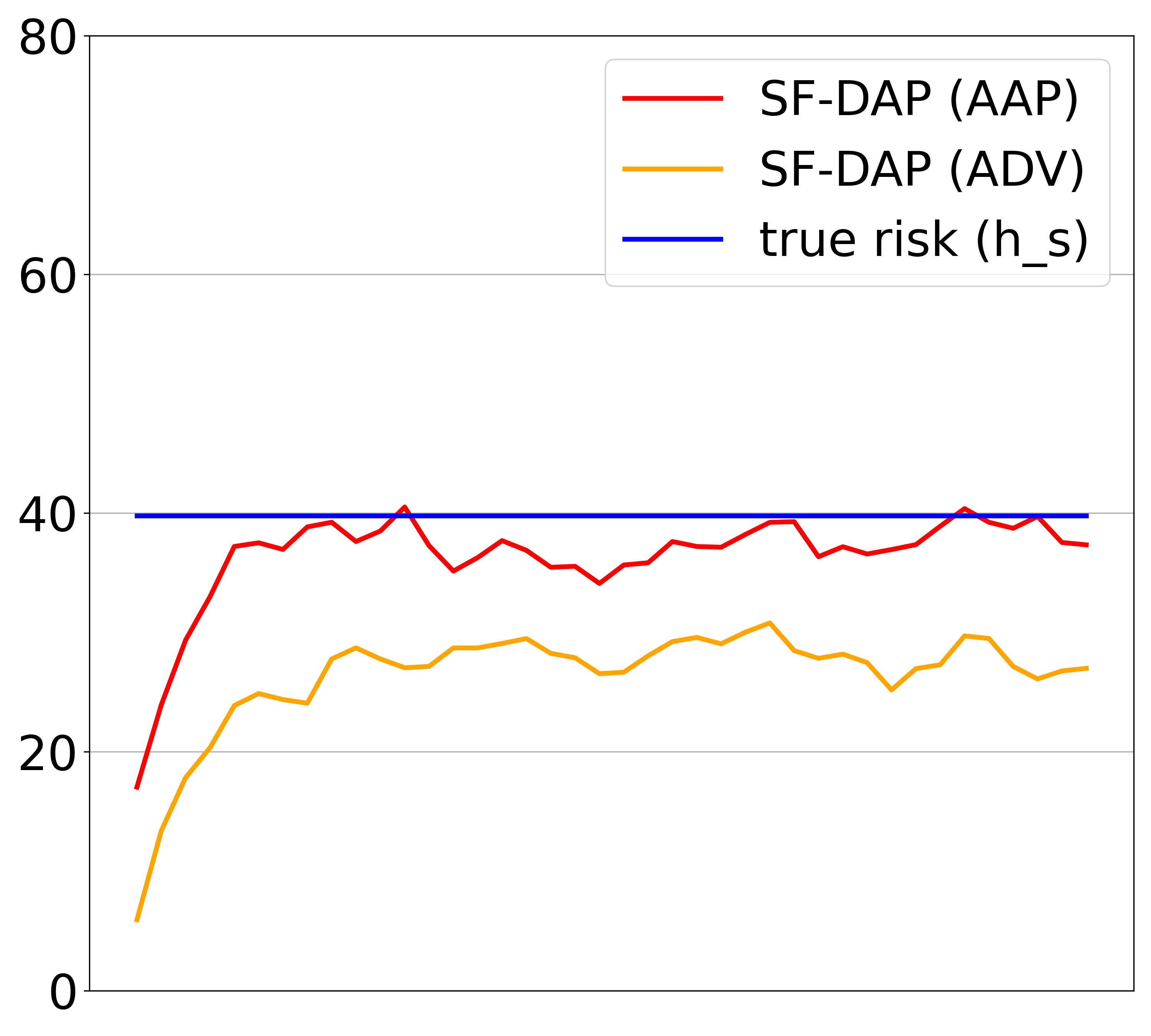}
  (e) Cl$\rightarrow$Pr
  \end{minipage}~
  \begin{minipage}[t]{0.28\textwidth} \centering \footnotesize
  \includegraphics[width=0.95\linewidth]{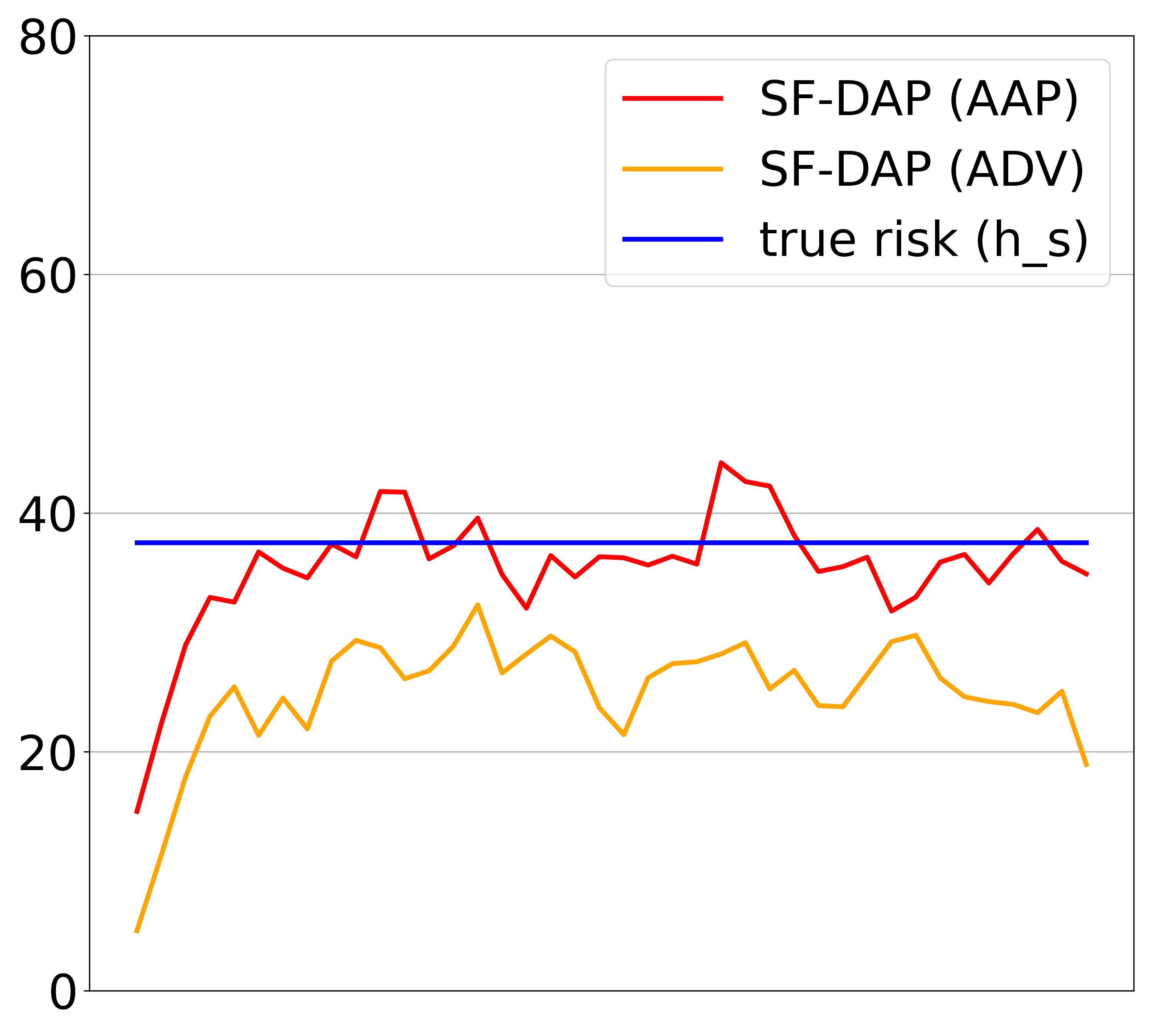}
  (f) Cl$\rightarrow$Re
  \end{minipage}\\
  \begin{minipage}[t]{0.28\textwidth} \centering \footnotesize
  \includegraphics[width=0.95\linewidth]{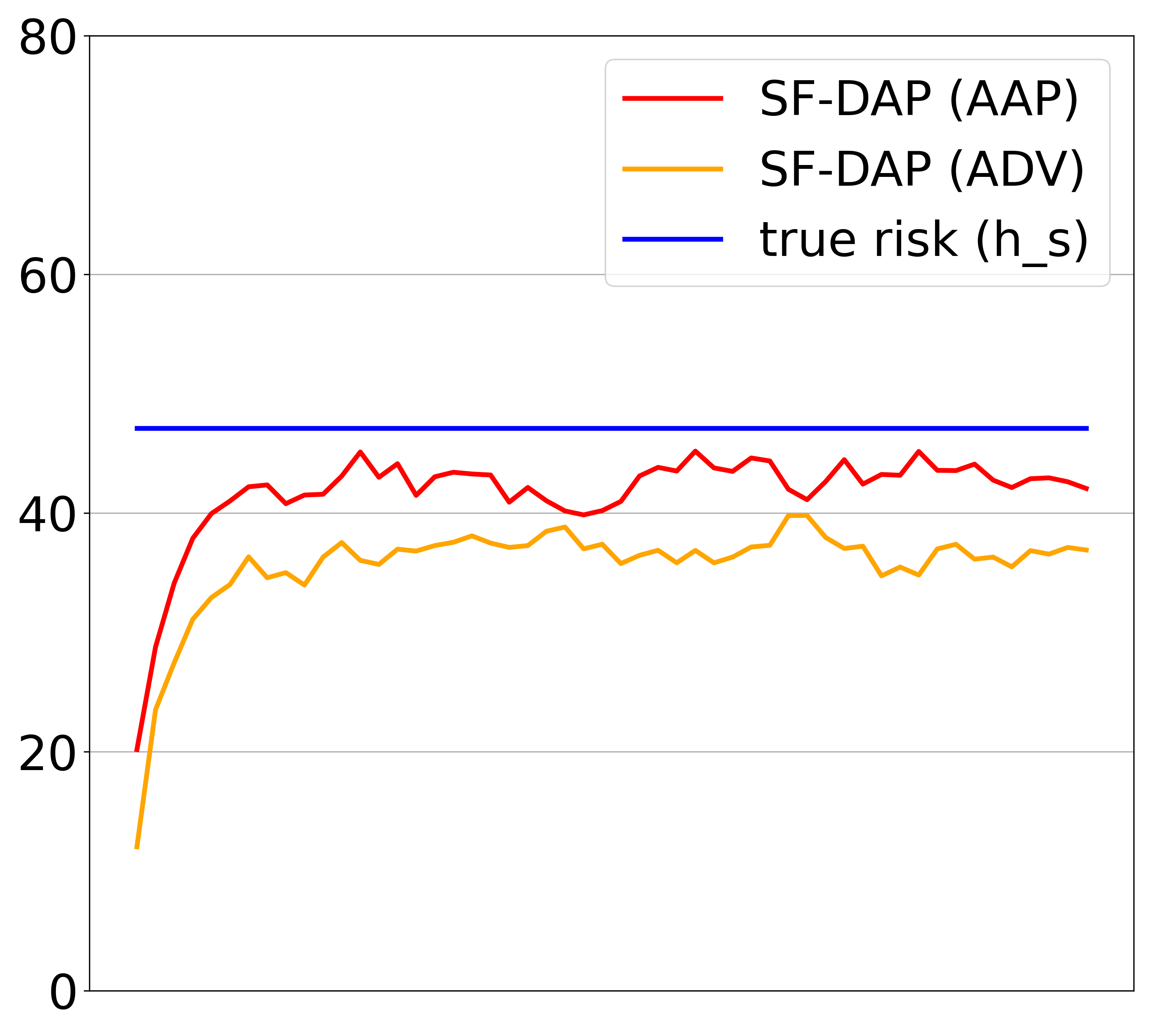}
  (g) Pr$\rightarrow$Ar
  \end{minipage}~
  \begin{minipage}[t]{0.28\textwidth} \centering \footnotesize
  \includegraphics[width=0.95\linewidth]{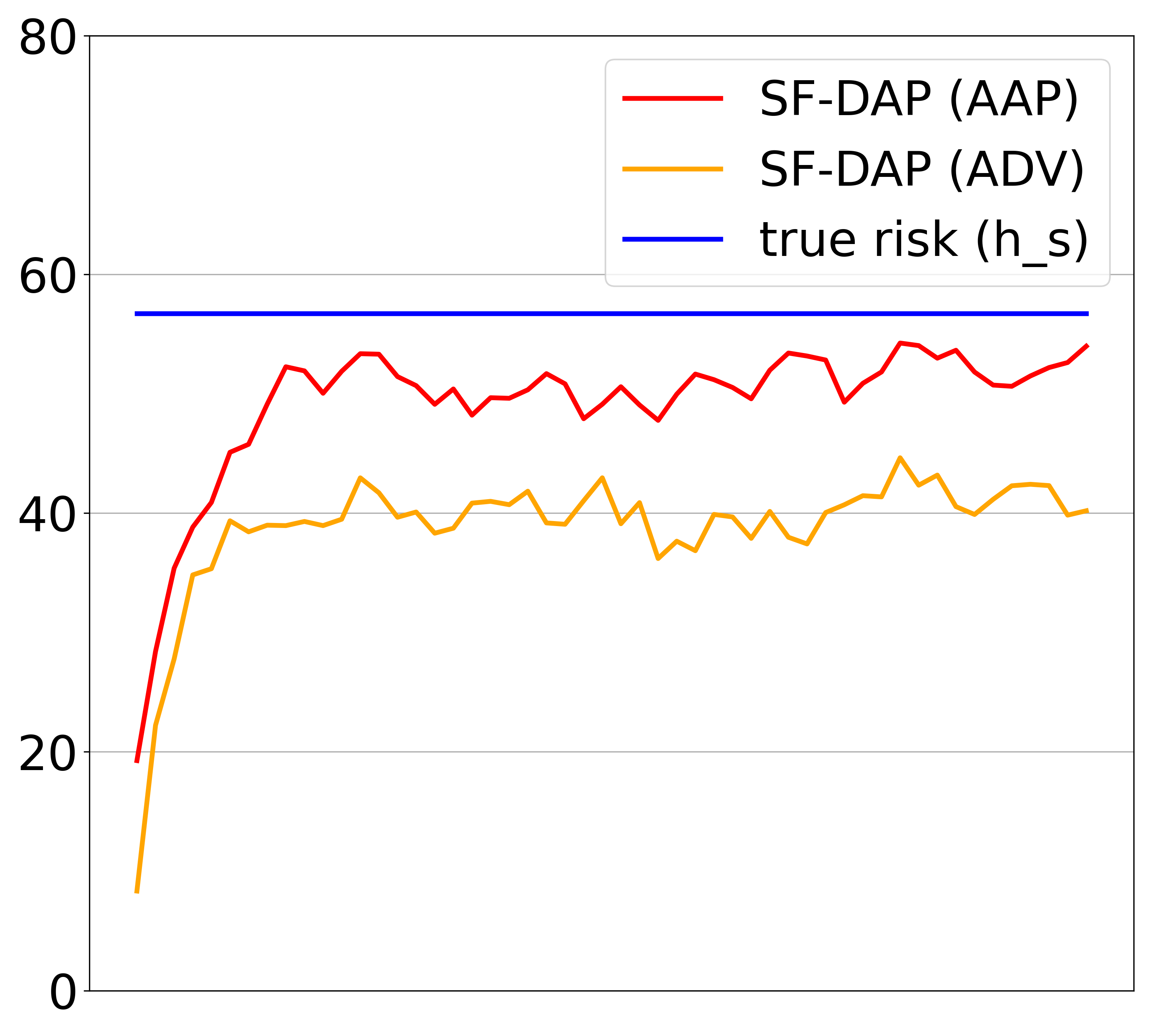}
  (h) Pr$\rightarrow$Cl
  \end{minipage}~
  \begin{minipage}[t]{0.28\textwidth} \centering \footnotesize
  \includegraphics[width=0.95\linewidth]{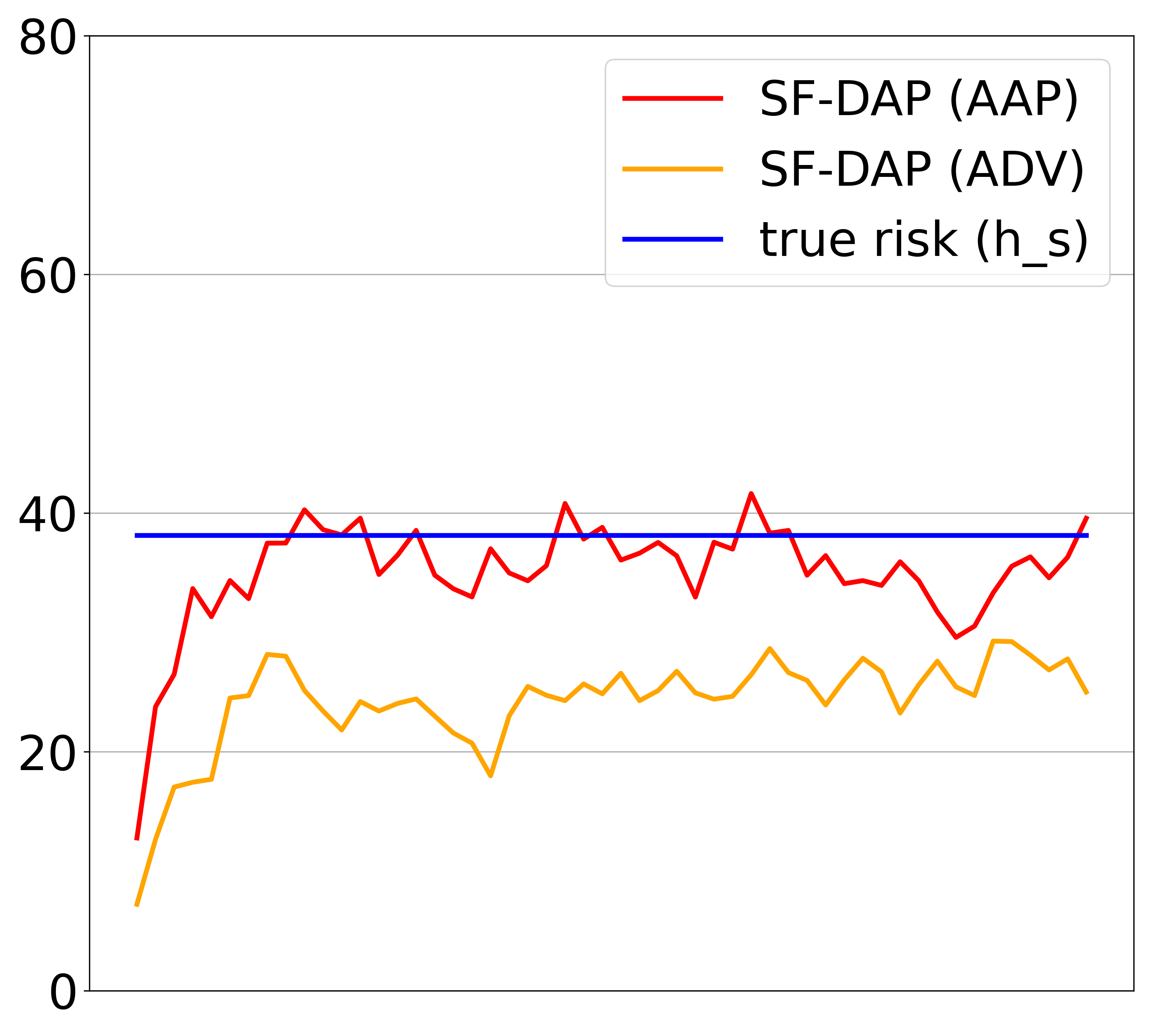}
  (i) Pr$\rightarrow$Re
  \end{minipage}\\
  \begin{minipage}[t]{0.28\textwidth} \centering \footnotesize
  \includegraphics[width=0.95\linewidth]{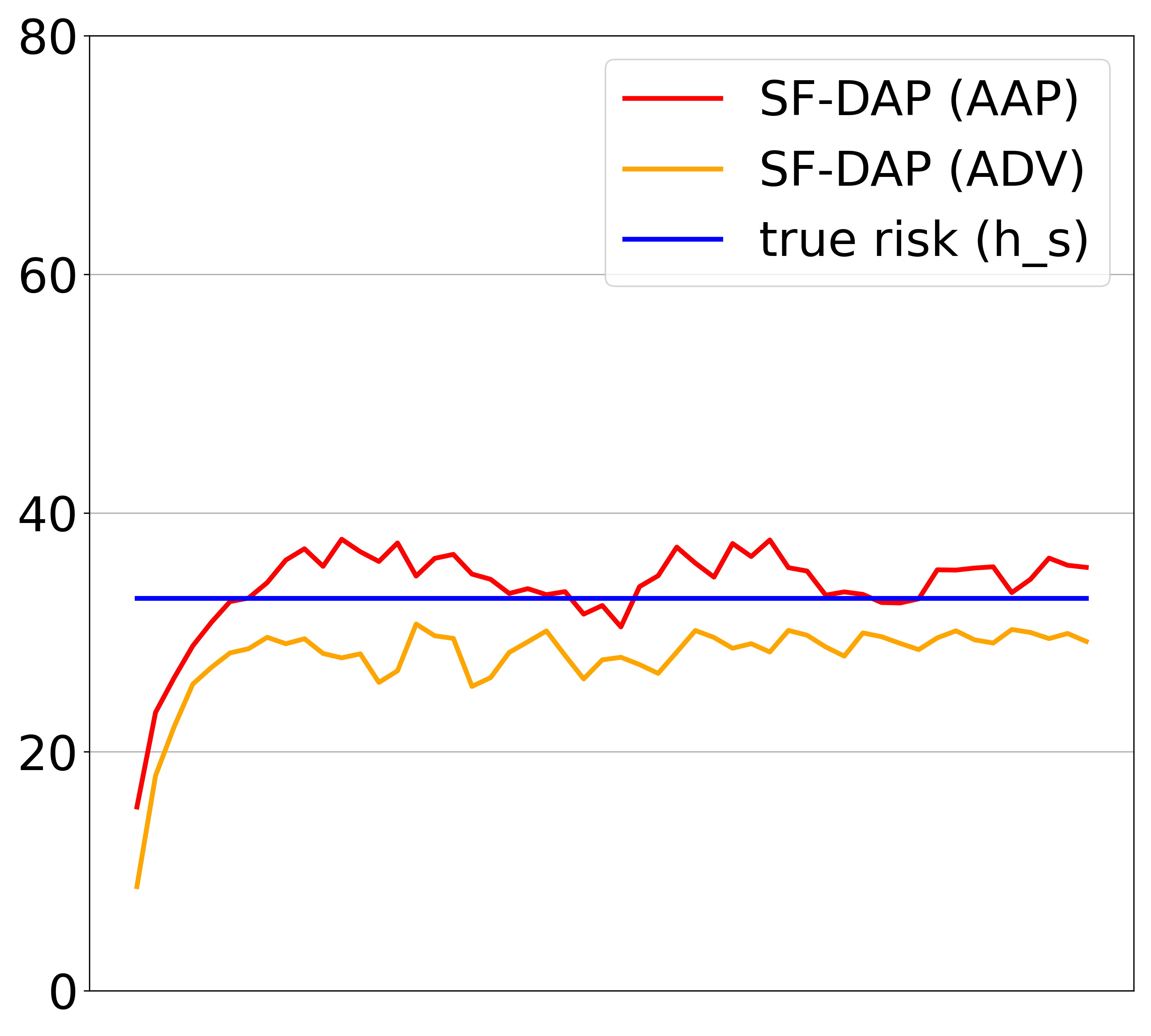}
  (j) Re$\rightarrow$Ar
  \end{minipage}~
  \begin{minipage}[t]{0.28\textwidth} \centering \footnotesize
  \includegraphics[width=0.95\linewidth]{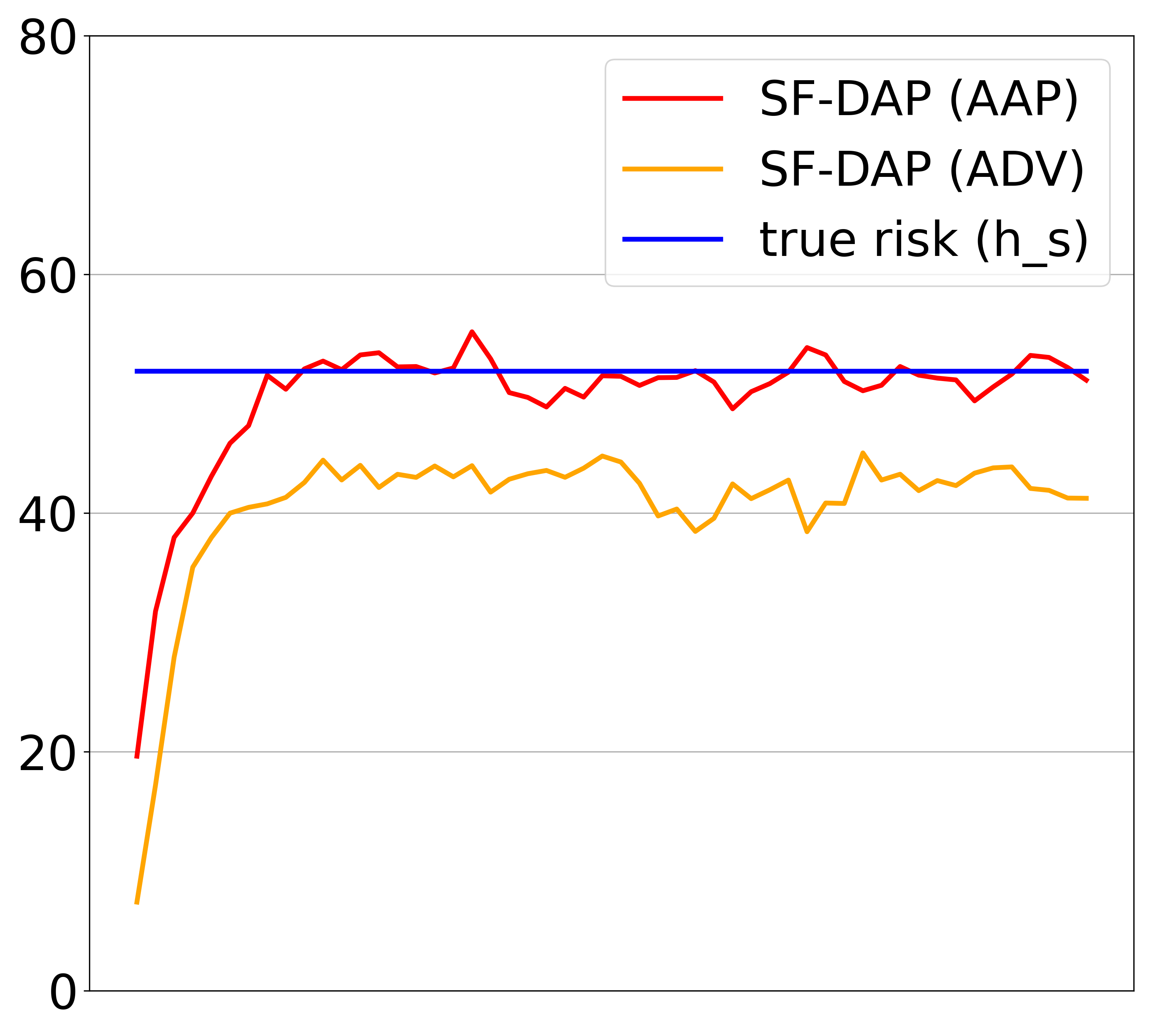}
  (k) Re$\rightarrow$Cl
  \end{minipage}~
  \begin{minipage}[t]{0.28\textwidth} \centering \footnotesize
  \includegraphics[width=0.95\linewidth]{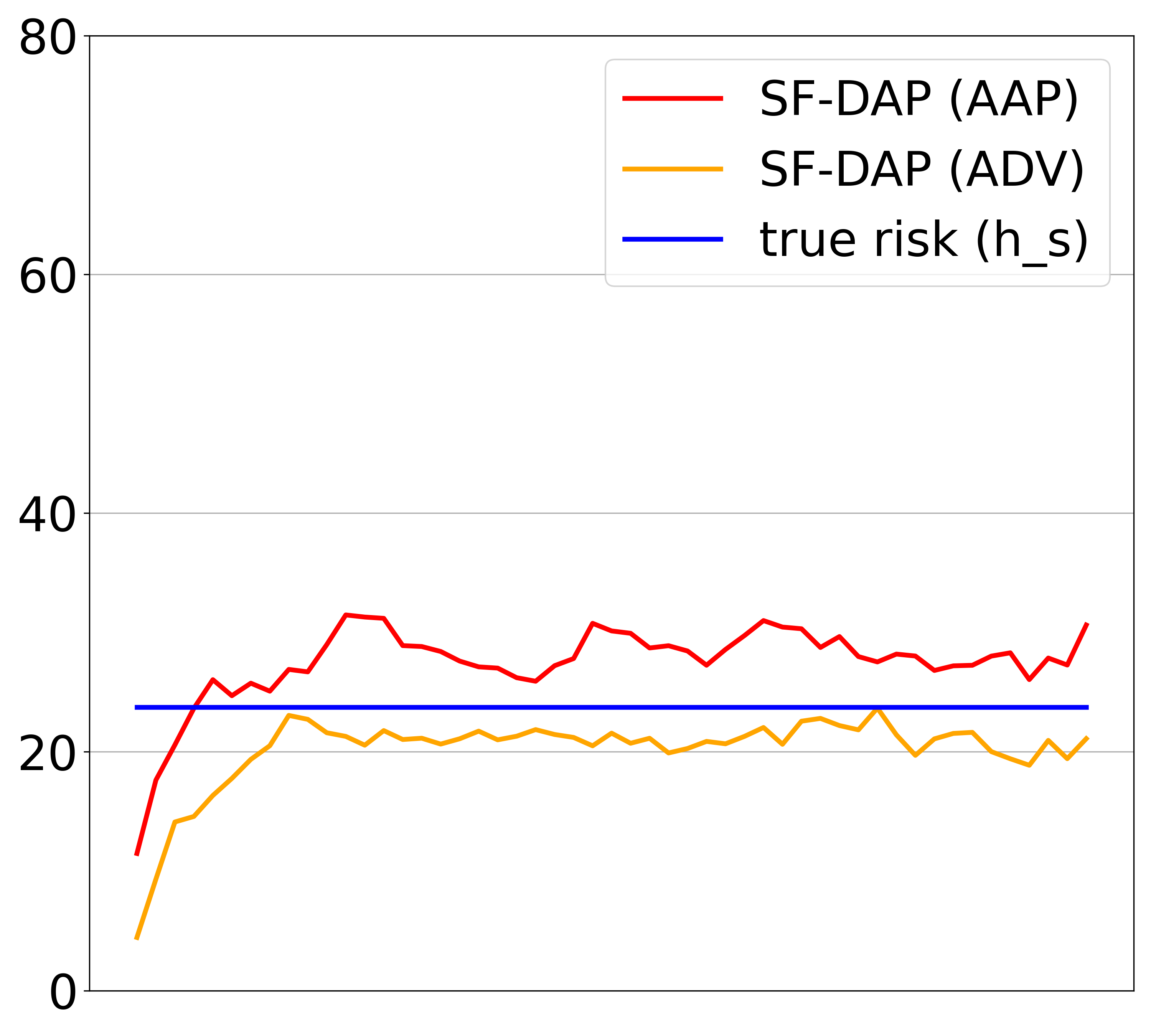}
  (l) Re$\rightarrow$Pr
  \end{minipage}\\
\vspace{0.5ex}
\caption{
(Best viewed in color) Performance trends of estimation as UDA progresses are presented for Office-Home benchmarks.
The true risk of the source model on the target data as well as the risk estimated by SF-DAP (AAP) and SF-DAP (ADV) are represented by blue, red and orange lines, respectively.
}
\label{fig:officehome_trend}
\end{figure*}

\begin{figure*}[!t]
  \centering
  \begin{minipage}[t]{0.28\textwidth} \centering \footnotesize
  \includegraphics[width=0.95\linewidth]{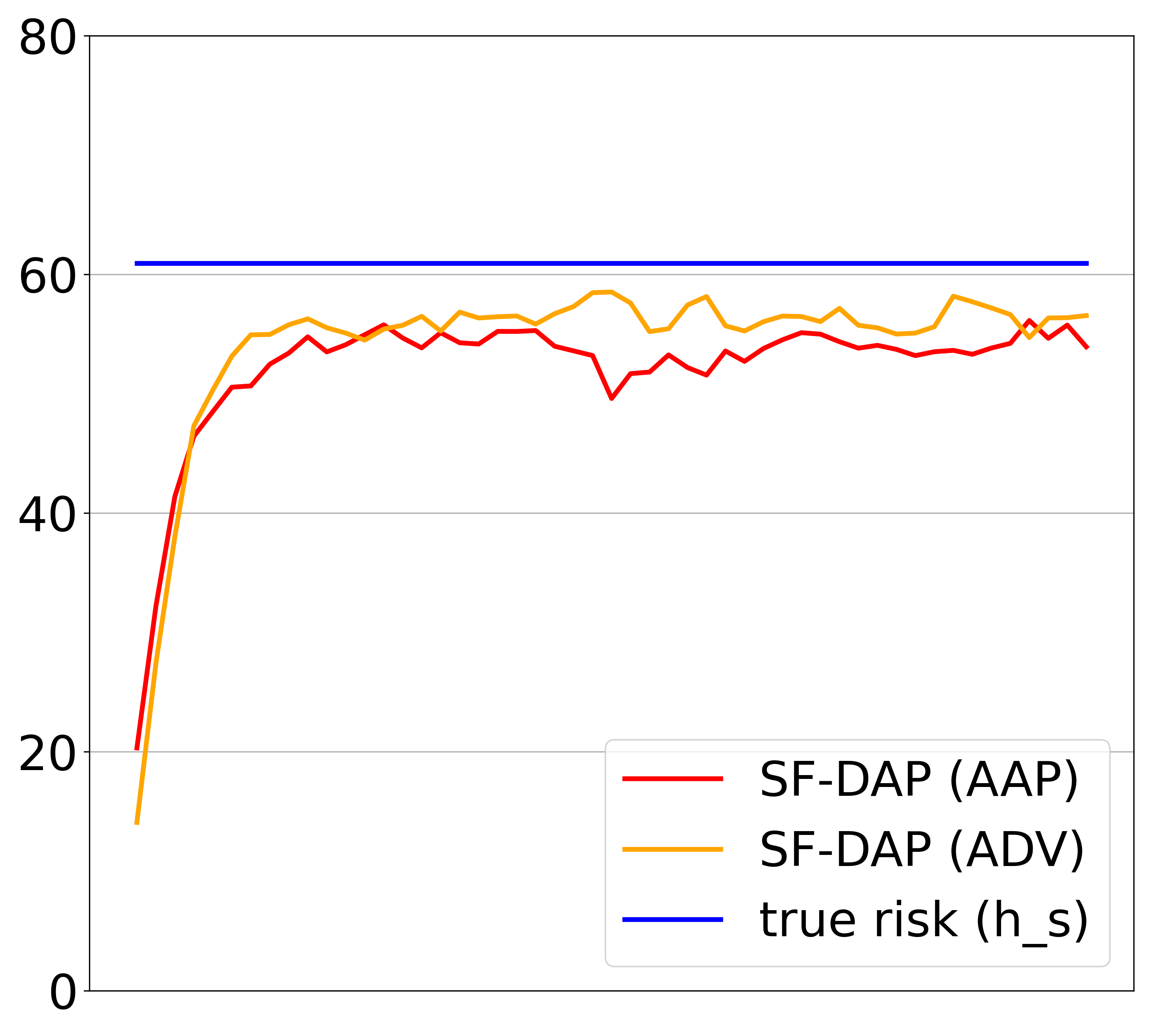}
  (a) MNIST$\rightarrow$SVHN
  \end{minipage}~
  \begin{minipage}[t]{0.28\textwidth} \centering \footnotesize
  \includegraphics[width=0.95\linewidth]{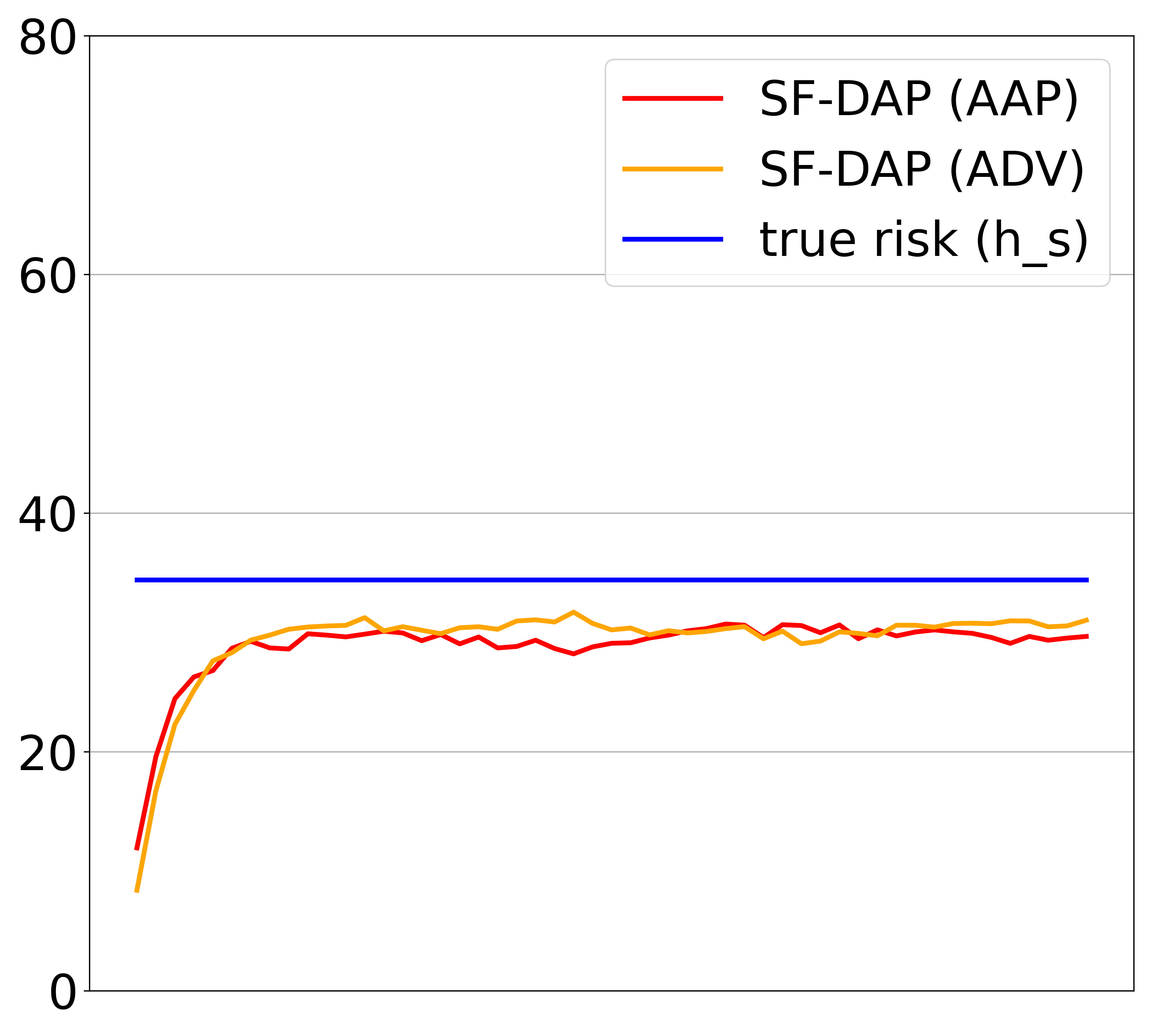}
  (b) USPS$\rightarrow$MNIST
  \end{minipage}~
  \begin{minipage}[t]{0.28\textwidth} \centering \footnotesize
  \includegraphics[width=0.95\linewidth]{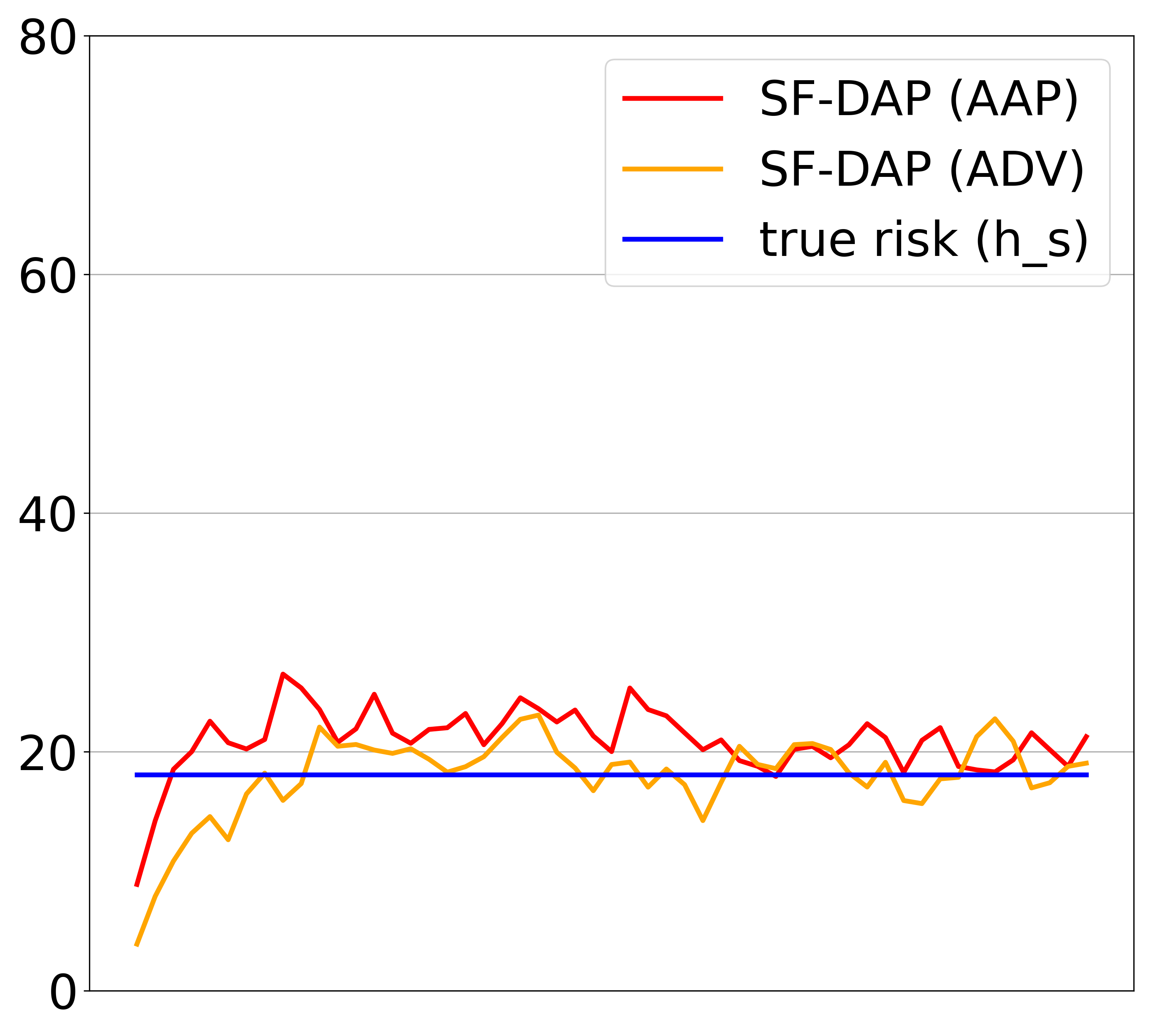}
  (c) Amazon$\rightarrow$DSLR
  \end{minipage}\\
  \begin{minipage}[t]{0.28\textwidth} \centering \footnotesize
  \includegraphics[width=0.95\linewidth]{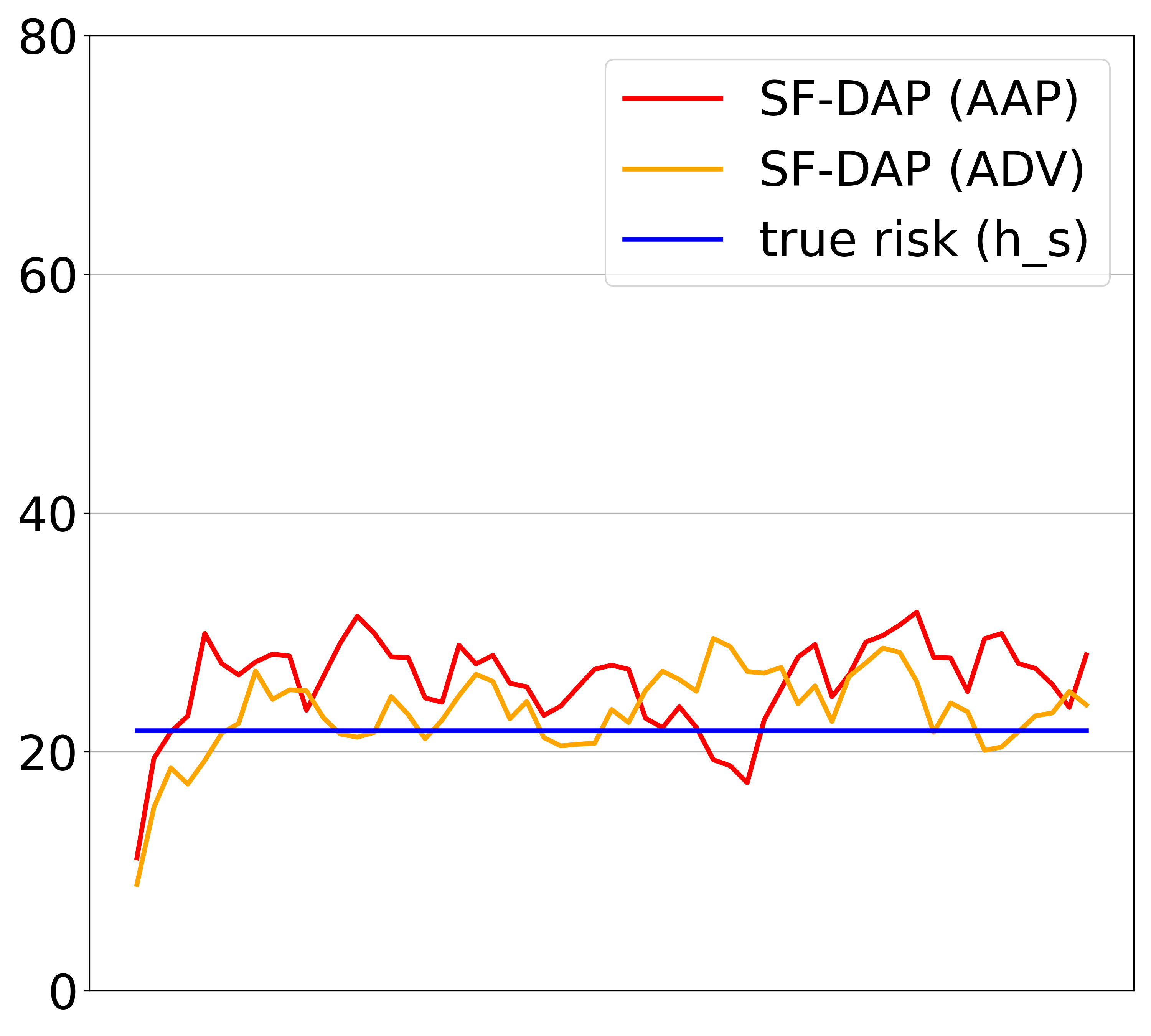}
  (d) Amazon$\rightarrow$Webcam
  \end{minipage}~
  \begin{minipage}[t]{0.28\textwidth} \centering \footnotesize
  \includegraphics[width=0.95\linewidth]{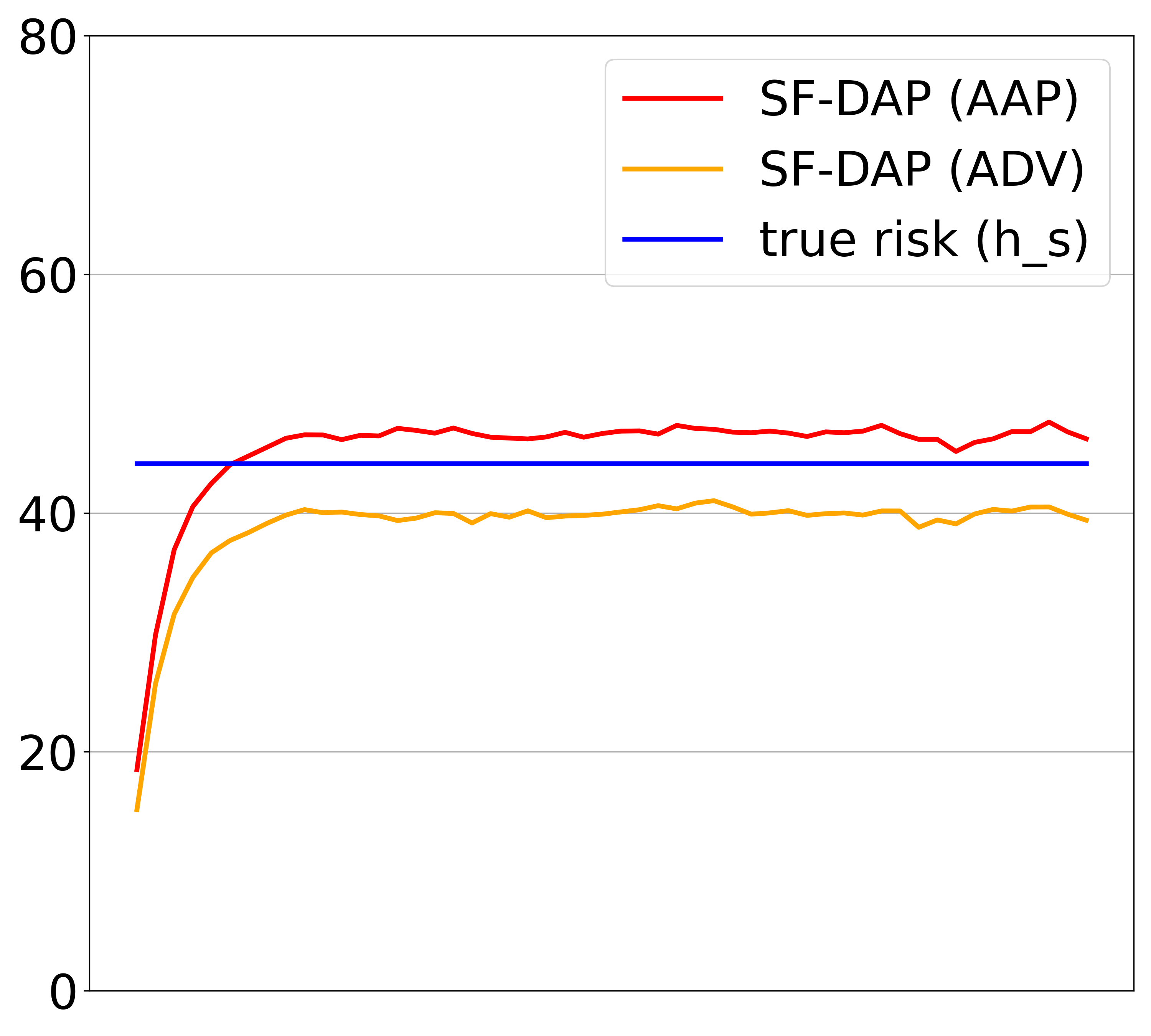}
  (e) VisDA
  \end{minipage}~
  \begin{minipage}[t]{0.28\textwidth} \centering \footnotesize
  \includegraphics[width=0.95\linewidth]{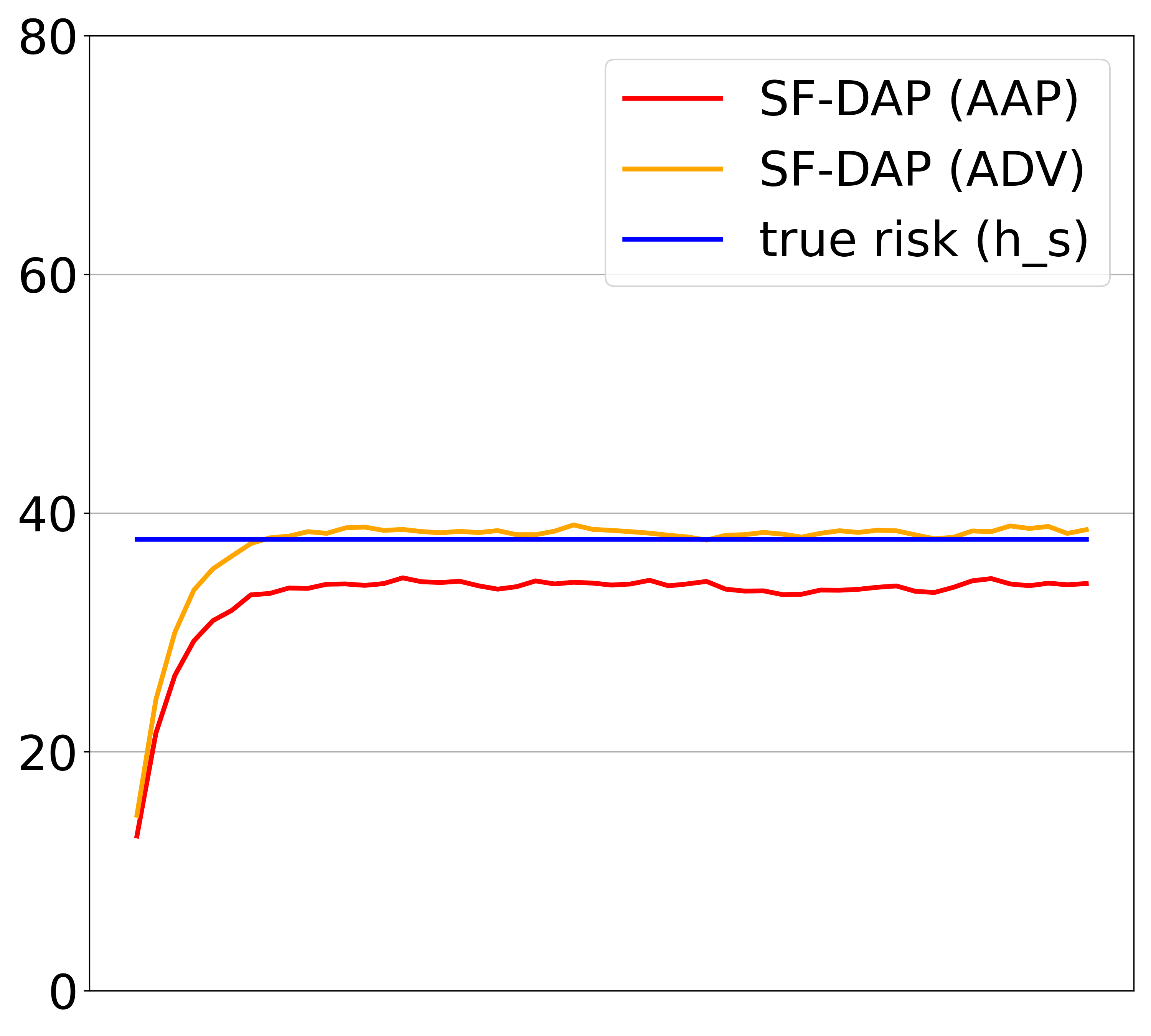}
  (f) Impulse noise
  \end{minipage}\\
  \begin{minipage}[t]{0.28\textwidth} \centering \footnotesize
  \includegraphics[width=0.95\linewidth]{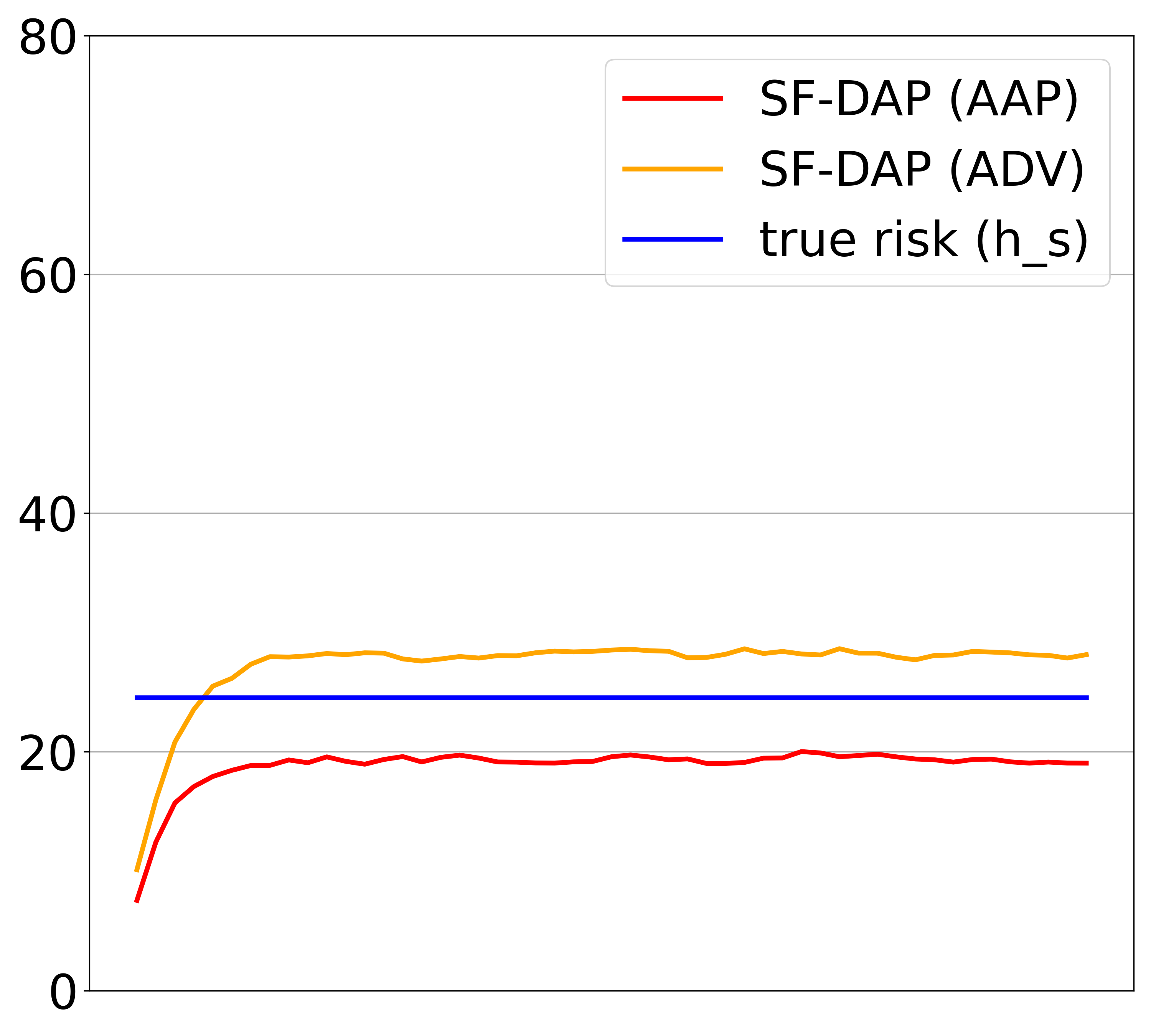}
  (g) Fog
  \end{minipage}~
  \begin{minipage}[t]{0.28\textwidth} \centering \footnotesize
  \includegraphics[width=0.95\linewidth]{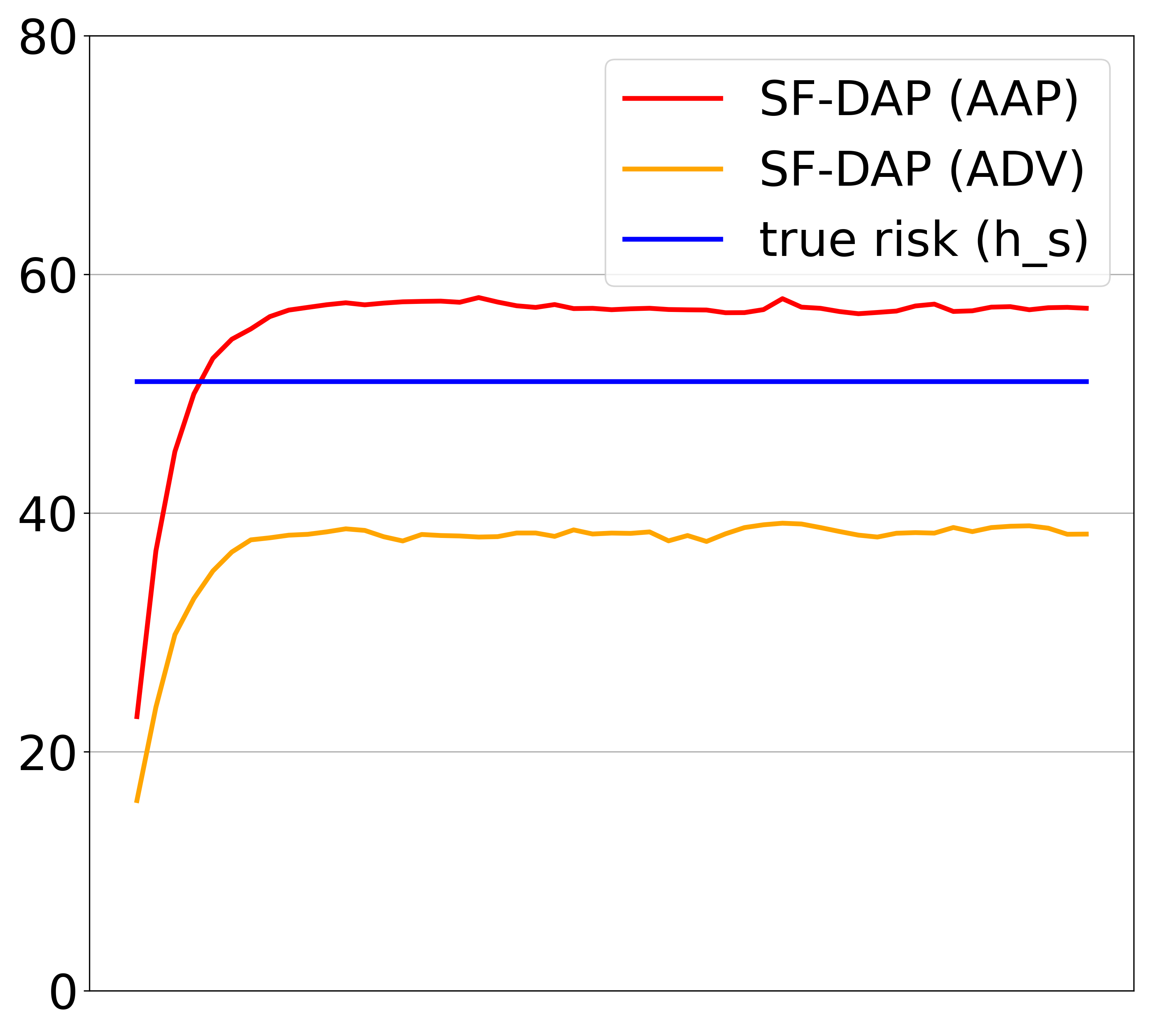}
  (h) Glass blur
  \end{minipage}~
  \begin{minipage}[t]{0.28\textwidth} \centering \footnotesize
  \includegraphics[width=0.95\linewidth]{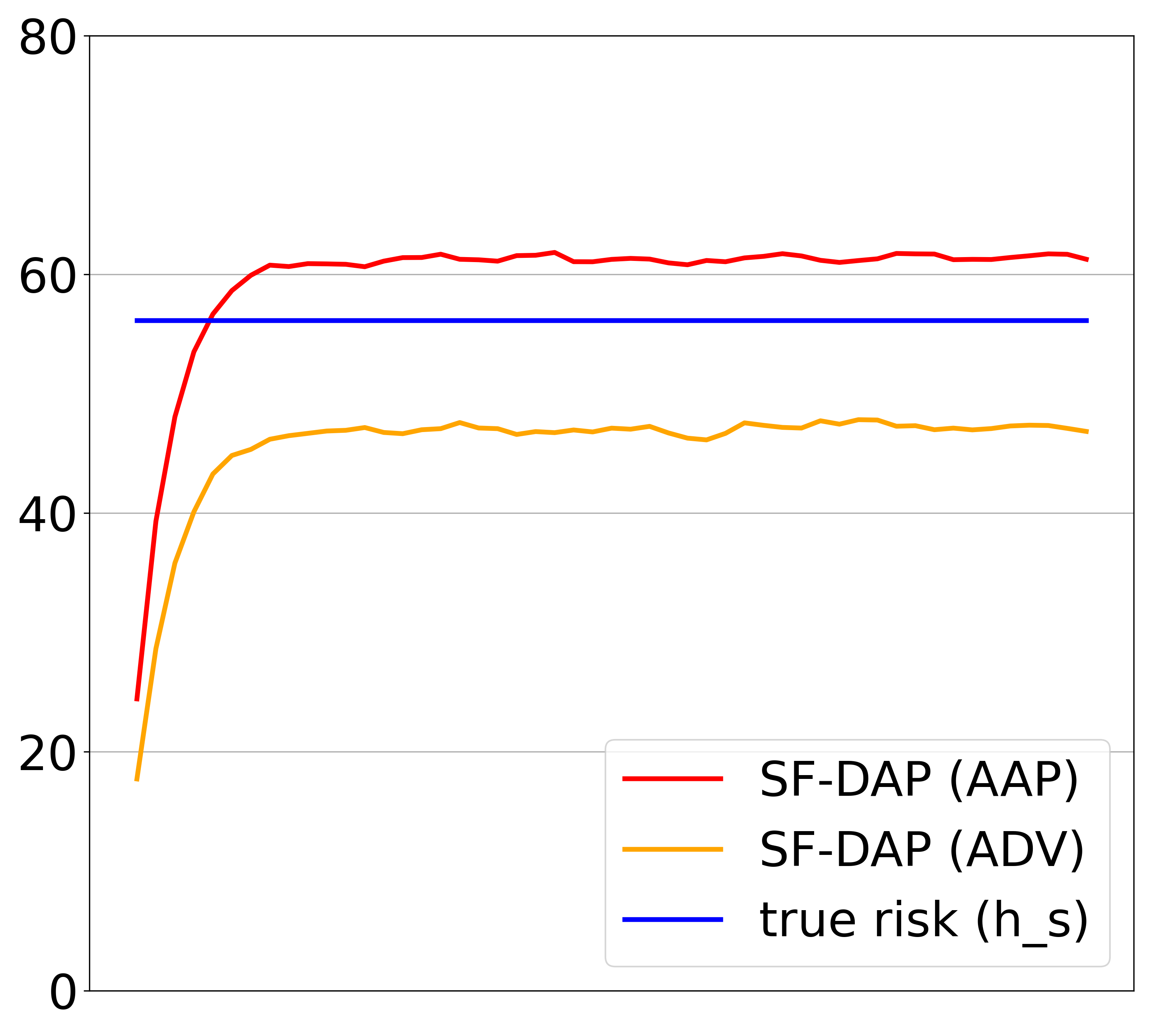}
  (i) Snow
  \end{minipage}\\
\vspace{0.5ex}
\caption{
(Best viewed in color) Performance trends of estimation as UDA progresses are presented for some of Digits, Office-31, CIFAR-10-C, CIFAR-100-C and VisDA benchmarks.
(f) and (g) depict some results in 19 CIFAR-10 $\rightarrow$ CIFAR-10-C experiments, whereas (h) and (i) display outcomes in CIFAR-100 $\rightarrow$ CIFAR-100-C experiments.
}
\label{fig:others_trend}
\end{figure*}

We track the performance trend of accuracy estimation as the proposed source-free UDA, PAFA, progresses to each target domain.
Our proposed framework, SF-DAP, starts producing accurate estimates surprisingly early and remains steady throughout the rest of the UDA iterations as illustrated in Fig \ref{fig:officehome_trend} and Fig \ref{fig:others_trend}.

%------------------------------------------------------------------------
%------------------------------------------------------------------------

%------------------------------------------------------------------------
%------------------------------------------------------------------------

\end{document}